\documentclass[times,twocolumn,final]{elsarticle}
\usepackage{cag}
\usepackage{framed,multirow}

\usepackage{amssymb}
\usepackage{latexsym}

\usepackage{amsmath}
\usepackage{graphicx}
\usepackage{amsfonts}
\usepackage{caption}

\usepackage{url}
\usepackage{xcolor}
\definecolor{newcolor}{rgb}{.8,.349,.1}

\usepackage{hyperref}

\usepackage[switch,pagewise]{lineno} %Required by command \linenumbers below

\journal{Computers \& Graphics}

\begin{document}

\verso{LPMNet: Latent Part Modification and Generation for 3D Point Clouds}

\begin{frontmatter}

\title{LPMNet: Latent Part Modification and Generation for 3D Point Clouds}%

\author[1]{Cihan \snm{Öngün}\corref{cor1}}
%\cortext[cor1]{Corresponding author: 
%  Tel.: +0-000-000-0000;  
%  fax: +0-000-000-0000;}
%\emailauthor{congun@metu.edu.tr}{Cihan Öngün}
\ead{congun@metu.edu.tr}
    
\author[2]{Alptekin \snm{Temizel}}%\fnref{fn1}}
%\fntext[fn1]{Footnote 1.}  
\ead{atemizel@metu.edu.tr}

\address{Graduate School of Informatics, Middle East Technical University, Ankara, Turkey}
%\address[2]{Address, City, Postcode, Country}

%\received{1 February 2017}
\received{\today}
%%%% Do not use the below for submitted manuscripts
%\finalform{28 March 2017}
%\accepted{2 April 2017}
%\availableonline{15 May 2017}
%\communicated{S. Sarkar}

\begin{abstract}
%%%
In this paper, we focus on latent modification and generation of 3D point cloud object models with respect to their semantic parts. Different to the existing methods which use separate networks for part generation and assembly, we propose a single end-to-end Autoencoder model that can handle generation and modification of both semantic parts, and global shapes. The proposed method supports part exchange between 3D point cloud models and composition by different parts to form new models by directly editing latent representations. This holistic approach does not need part-based training to learn part representations and does not introduce any extra loss besides the standard reconstruction loss. The experiments demonstrate the robustness of the proposed method with different object categories and varying number of points. The method can generate new models  by integration of generative models such as GANs and VAEs and can work with unannotated point clouds by integration of a segmentation module. %The code is publicly available at \url{https://github.com/cihanongun/LPMNet}.
%%%%
\end{abstract}


\end{frontmatter}

\section{Introduction}
\label{sec:introduction}
Deep learning applications in the 3D domain are becoming increasingly more popular, expanding on the already successful applications in the 2D image domain and there is a surge in the number of studies focusing on the artificial generation of 3D models. Artificially generated 3D models have many uses in virtual environments, simulations, and 3D printing. Leading companies are now providing AI tools that help users create better 3D models, make recommendations for more realistic models and correct errors in graphics for a better user experience.   

A number of different data types can be used to represent 3D models. While mesh-based representation is popular in computer graphics, voxel-based representation is preferred in 3D data processing applications because of its simplicity. On the other hand, point clouds are the most prominent data type in 3D perception of the real world and they are popular in various fields such as 3D scanners, robotics, autonomous cars, face recognition, and human pose estimation. Detection, recognition and segmentation are the main tasks in these fields and generation of 3D models in point clouds is expected to facilitate new types of approaches for these tasks.

Real-world objects are composed of individual parts and model generation systems should ideally be part-aware in-line with this semantic composition. The basic approach in the literature is to generate parts separately and then assemble them to form the complete object. However, this approach needs training different networks which are experts on specific parts and a separate network to combine these parts. In this paper, we propose a holistic approach to learn the semantic properties of the parts with a single neural network model. The proposed architecture is an Encoder-Decoder network that represents the parts, in addition to the global shape, separately in the feature space. Making modifications in the feature space allows meaningful modifications by preserving semantic properties. This is in contrast to the traditional way of making modifications in the input space which results in a completely new model. The contributions of the proposed method are as follows:
\begin{itemize}
    \item It handles part editing,  modification and global model generation with a single architecture and eliminates the need for an additional network for part assembly. The parts generated by modifications of latent space stay coherent with the global shape.
    \item It does not require any additional loss function other than the standard reconstruction loss.
    \item It provides a generic solution to convert regular generative networks based on PointNet feature extraction into part-aware networks.
    \item It is scalable and can be used with different point cloud sizes, objects having different numbers of parts and parts having different resolutions.
    \item It can process models without any explicit part information during inference by integration of a segmentation module.
\end{itemize}

The paper is structured as follows: Section \ref{sec:relatedWork} summarizes the literature on point cloud generation with necessary background information. Section \ref{sec:proposedMethod} explains the proposed method in detail. Section \ref{sec:experimentalEvaluation} gives the details of the experiments and the visualization of sample results. Section \ref{sec:conclusions} provides the conclusions and directions for future work.  

\section{Background and Related Work}
\label{sec:relatedWork}
\subsection{Point clouds}
\label{sec:pointClouds}
Point clouds are a set of unstructured points in a 3D coordinate system that defines 3D models. Capturing, visualizing and modification of point clouds are simpler compared to the other 3D representation methods since the data points only have position variables for a point \(p\) and some extra information such as color value when needed. A 3D model can be defined by a varying number of points and the higher the number points, the better and more detailed is the representation. While capturing and modification of point clouds is straightforward, the processing in this domain is challenging due to the following properties:

\textit{Point clouds are unstructured and points have no connectivity information.} The nearest or sequential points cannot be assumed to be neighbors since they may be in different semantic parts. The proposed method uses a point-wise feature extractor to process points independently without any connection information.

\textit{Points in a point cloud model can be in any order.} A point cloud with $N$ points can be defined by $N!$ permutations of ordering. The proposed method uses order invariant part and global feature extractors to deal with the ordering problem.

\textit{Point clouds can have arbitrary number of points.} The number of points is not constant and can be increased or decreased to have different resolutions. However, most of the models assume a fixed input size. The proposed method utilizes max-pooling operation to extract the important points for feature extraction allowing use of an arbitrary number of points.

PointNet \cite{qi2016pointnet} is the most popular neural network based approach for point cloud processing. It provides an end-to-end solution to extract global and local features and it is an effective baseline for a range of tasks such as object classification, part segmentation, and scene semantic parsing. PointNet++ \cite{qi2017pointnetplusplus} is an extended version of the original PointNet which uses a hierarchical neural network that applies PointNet recursively on a nested partitioning of the input point set. PointNet++ uses sampling and grouping layers to extract features from local point neighborhoods. Neighboring points may belong to different parts, so these layers must also be redesigned for part considerations. As the proposed method introduces a new step for part feature extraction in intermediate layers, it would not be possible to use PointNet++ directly. Hence the standard PointNet is adopted since it provides a holistic approach for feature extraction.

Some approaches convert point clouds into different representations to tackle with the aforementioned problems. DeepSDF \cite{park2019deepsdf} uses Signed Distance Functions to represent 3D shapes with continuous functions for easier processing of them in neural networks. While continuous functions do not suffer from the same problems as point clouds, pre-processing and post-processing steps are necessary for conversion. Also, it is not straightforward to represent semantic parts of 3D shapes with continuous functions. PointConv \cite{pointconv}, KPConv \cite{kpconv}, VV-Net \cite{vvnet} and Monte Carlo Convolution \cite{montecarlo} focus on developing new convolutional methods. While these studies are reported to have better results than PointNet for segmentation and classification, they are not designed for point specific feature extraction. Hence, they are not directly applicable for the part modification and generation problems, which are the main objectives of this paper.

\subsection{Generative Models}
\label{sec:generativeModels}
\paragraph{Generative Adversarial Networks (GAN)} \cite{goodfellow2014generative} consist of 2 different neural networks; Generator \textit{G} and Discriminator \textit{D}. While the Generator generates new realistic samples, Discriminator aims to distinguish between real and fake samples and it is trained by a loss measure calculating the difference between the predictions and true values. Generator aims to fool the Discriminator so it needs to generate as realistic samples as possible. At each iteration, Discriminator gets better at distinguishing real and fakes samples and Generator gets better at generating more realistic samples. The whole system is a minimax game between Generator and Discriminator. Assuming $x$ is real data and $z$ is a latent variable, GAN loss function can be defined as:
\begin{equation}
\begin{split}
\underset{G}{min}\,\underset{D}{max}V(D,G)\; = \;&E_{x\sim p_{data}(x)}[log~D(x)] \;+ \\ &E_{z\sim p_{z}(z)}[log~ (1-D(G(z)))]
\end{split}
\end{equation}
While GAN can generate novel and realistic samples, training may become unstable in the long run, resulting in mode collapse. Also GAN suffers from lack of diversity in generated samples. WGAN \cite{arjovsky2017wasserstein} proposes a better objective function using Wasserstein distance to address these problems:
\begin{equation}
\begin{split}
\underset{G}{min}\,\underset{D}{max}V(D,G)\; = \;&E_{x\sim p_{data}(x)}[D(x)] \;- \\ &E_{z\sim p_{z}(z)}[D(G(z))]
\end{split}
\end{equation}

\paragraph{Variational Autoencoder (VAE)} \cite{kingma2013auto} architecture is an extension of Autoencoder (AE) architecture addressing the content generation problem and the main difference lies in the bottleneck layer. AEs represent each input sample with a latent variable in a lower dimension. This may lead to an overfitting problem since the network is not trained for a regularized latent space. Latent space may not be continuous and some points in this latent space may represent meaningless samples in the input space. VAEs represent each input sample with a distribution by adding a regularization loss to the reconstruction loss. Regularization imposes latent space to belong to a standard normal distribution so any random point generates a new meaningful sample.

A comprehensive analysis of different point cloud generation models is provided in \cite{achlioptas2017latent_pc} where the PointNet model is used as an Encoder and a multi-layer perceptron is used as a Decoder. Chamfer Distance (CD) and Earth Mover’s Distance (EMD) are used to calculate the reconstruction loss. PointFlow \cite{pointflow} proposes a probabilistic framework for 3D point cloud generation using continuous normalizing flows. To modify the generated samples of these global shape generators, interpolation and latent space arithmetic are used. While these techniques can be used to modify samples generated by all different latent representation models (AEs, GANs, etc.), they only allow control over the existence of an attribute and not the desired shape. Also, direct part modification is not possible since there is only a global latent code that controls the shape with an entangled representation.

For part editing and generation, the most popular approach is reconstructing or generating the parts separately by different networks and then assembling them to form the global shape by an additional composition network. In \cite{dubrovina2019composite}, a "Spatial Transformer Network" is used to combine the generated parts by applying affine transformations. CompoNet \cite{Schor_2019_ICCV} uses a separate Encoder-Decoder model for each part. Encoders are used to get codes for each part and a composition network outputs transformation parameters per part. The generated parts are warped together using the transformation parameters. In \cite{li2019learning}, VAE-GANs (Variational Autoencoder Generative Adversarial Networks) are used to generate parts instead of naive AEs. VAE-GAN uses a Variational Autoencoder instead of a Generative network, so it is an Encoder-Decoder-Discriminator architecture. In \cite{G2L18}, an inverse approach is adopted where a low-resolution global shape is generated first and then a part refiner module enhances the generated parts by refining and completing the missing regions. Most of these studies use voxels as input data because of the ease of data processing. Most part based studies assume that different parts have the same number of points. Tree-GAN \cite{treegan} uses a tree-structured graph convolution network for multi-class generation. It allows semantic part generation and modification of newly generated samples, but lacks the ability to encode and reconstruct existing shapes.

StructureNet \cite{Mo2019StructureNetHG} (followed by StructEdit \cite{structedit}) is one of the pioneer studies for part editing and generation. It uses two encoders and two decoders, one to process geometry and one to process relations between parts with graph networks. The main aim is to encode-decode structures as well as generating new ones. While the results are very detailed, the model requires training with fine-grained and hierarchical part annotations, which is not always available. We designed our system to work with a simple labeling indicating to which part a point belongs to. Also we expect from our system to learn the relations between parts without specifically trained for it since it operates on latent space for semantic modifications.

There are a few studies that directly operates on meshes. SDM-Net \cite{sdmnet} generates structured deformable meshes using a 2-level VAE based approach for learning part geometries and structures. COALESCE \cite{coalesce} aims for component-based shape assembly to align the parts and synthesize part connections to form plausible shapes. It uses two different networks for alignment and joint synthesis tasks.

The studies in the literature use multiple neural networks with different architectures to solve the problem of shape generation with respect to parts. The parts are generated independently and then they are processed by scaling, positioning and rotating to form a meaningful global shape. We aim to solve the problem with a single neural network that can handle part-aware global shape generation without any need for additional processing to form a meaningful global shape. The disentangled latent space allows exchanging and removal of existent parts or generation of new parts that fits the global model. Part generation is an intermediate step of the main process that results in global shape generation. The proposed method provides a holistic approach that generates the global shape with respect to part semantics instead of generating the parts separately. The proposed method can work on unannotated point clouds with the additional segmentation ability. The simplicity of the approach allows using a smaller model with fewer parameters than previous studies. 

\section{Proposed Method}
\label{sec:proposedMethod}
The proposed method is an end-to-end system consisting of 3 modules: Feature extractor, Segmentation and Decoder which are explained in Sections \ref{sec:feature}, \ref{sec:segmentation}, and \ref{sec:decoder} respectively. A generative module can also be integrated to provide generative capabilities which is explained in Section \ref{sec:generativeCapabilities}.

\subsection{Feature Extractor}
\label{sec:feature}
\begin{figure*}
\centering
\includegraphics[width=\textwidth]{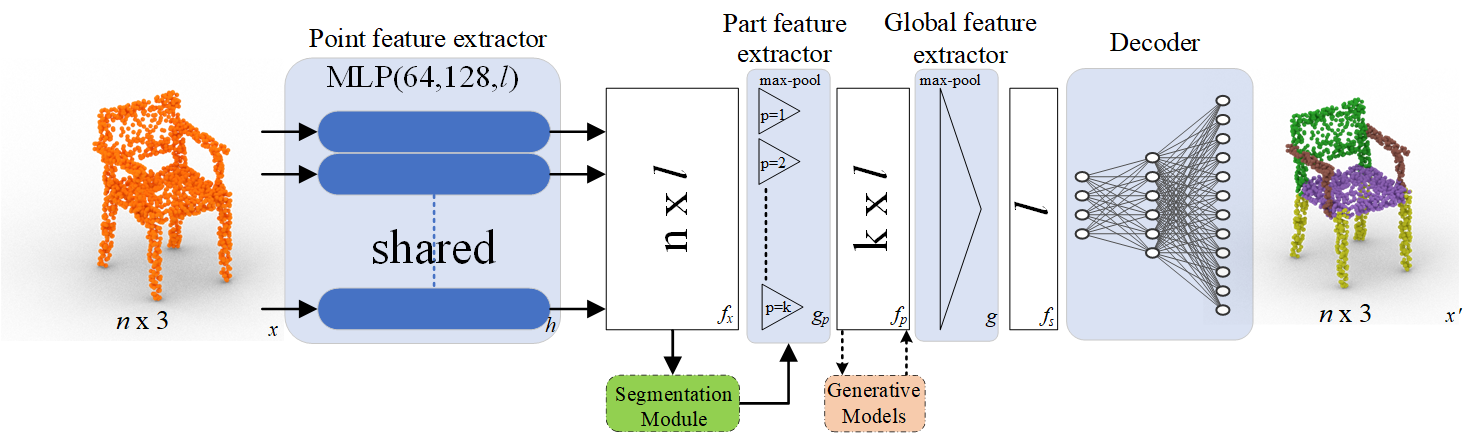}
\caption{The proposed architecture consists of a point-wise feature extractor, a part feature extractor, a global feature extractor and a decoder. The optional generative model allows generation of new parts and models. The optional segmentation module allows the system to work with unlabeled data.}
\label{fig:proposed}
\end{figure*}

The feature extractor is based on a modification of the standard PointNet architecture and introduces a part feature extraction step between the point feature extraction module and the global symmetric function (Fig. \ref{fig:proposed}). The point feature extractor is a multi-layer perceptron (MLP) model that takes \textit{n} points and outputs \textit{l} features for each point. PointNet applies max-pooling on the first axis to get the global feature. Max-pooling is a symmetric function and it gives the same result for the same input in any order so it is invariant to permutations of the input set. In the proposed method, instead of directly applying a global max-pooling, max-pooling is applied on a part to get an individual part feature. After this step, max-pooling is applied again on these part features to obtain the global feature for the whole shape. The idea is based on a 2-stage max-pooling operation which can be defined as max of maxes similar to the "reduce max" operation in parallel programming. Directly applying max operation on a vector of numbers gives the same result as applying the operation in multiple iterations. In this context, the first max operation is used to get part features and the subsequent max is used to get the global feature. In this way, while obtaining the same global feature as the original network, a number of separate part features are also obtained. This operation is shown in Eq. \ref{Eq:partbased} where $h$ is approximated by MLP and symmetric function $g$ is max-pooling.

\begin{equation}
\label{Eq:partbased}
\begin{split}
f_{p=1,...,k}(\{x_1,...,x_n\}) &\approx g_{p=1,...,k}(h(x_1),...,h(x_n))\\
f_s(\{x_1,...,x_n\}) &= g(f_{p=1},...,f_{p=k})\\
f:2^{\mathbb{R}^n} \rightarrow \mathbb{R},  h:\mathbb{R}^n &\rightarrow \mathbb{R}^l, g:\mathbb{R}^l\times...\times \mathbb{R}^l \rightarrow \mathbb{R} 
\end{split}
\end{equation}

Assuming $S\in \mathbb{R}^{n \times 3} $ is a point cloud having \textit{n} points, the point feature extractor extracts $l$ features from each point $x$ outputting a $n\times l$ point feature matrix $f_x \in \mathbb{R}^{n \times l} $. Both part feature extractor and segmentation module are fed with the point feature matrix. The part labels are extracted by the segmentation module. The part feature extractor applies max-pooling on each part separately ($g_{p=1,...,k}$), by taking the part labels into account and produces $k$ separate part feature vectors ($f_p \in \mathbb{R}^{k \times l}$ ), each having a size of $l$. Then a $k\times l$ matrix is formed by concatenating these vectors together. The global feature extractor applies global max-pooling $g$ to produce a global feature $f_{s} \in \mathbb{R}^{l}$ of size $l$. By this way, $k$ individual part features, in addition to a global feature, are obtained. The part features can be modified individually to change the part only or the global feature can be modified to change the global shape. This allows modification of specific parts, in addition to the modification of global shapes.

\subsection{Segmentation}
\label{sec:segmentation}
The part feature extractor needs part labels to generate part features. In part-segmented point cloud datasets, for a model with $k$ parts (For example, a chair model has $k=4$ semantic parts; seat, back, arm and leg), represented with $n$ points, there are $n$ labels, associating each point with a part label. While there are part labels in annotated datasets, such information is rarely available in real conditions. Segmentation module is employed to segment the unlabeled point clouds to get part labels. It uses point features generated by the point feature extractor to generate per-point part labels. Then these labels are fed to the part feature extractor. During the training, the segmentation module is trained together with the system using the ground truth part labels from the training data. During inference, the segmentation module generates the part labels, eliminating the need for ground-truth part labels and making the system an end-to-end solution for unannotated point clouds.

As an alternative to end-to-end training with the whole system, the module can be trained in isolation or can be trained using a pretrained point-wise feature extractor. All training options generate similar results within a range of 2\% with respect to segmentation performance. The point features can be concatenated with global features to improve the segmentation performance, allowing segmentation by considering local and global features together. This method decreases segmentation loss significantly over using the point features only. The global features are extracted by a max operation on point features.

The aim of the segmentation module is to predict part labels when they are not available. If the part labels are available, then this module can be omitted and these labels can directly be fed into the part feature extractor. This makes the reconstruction performance better as expected since the part labels are not predictions but ground truths. While this is a better option for reconstruction performance, it eliminates the ability of the system to work with unannotated raw point clouds.

\subsection{Decoder}
\label{sec:decoder}
The aim of the decoder is to generate a $n\times3$ point cloud from the global feature vector $l$. An MLP or a Deconvolutional model can be employed for this purpose. The decoder is trained with reconstruction loss to enforce reconstruction of a given sample with the minimum loss. Decoder learns to generate corresponding global shapes for given global feature vectors. Modified feature vectors are fed to the decoder to get the modified point cloud models. Segmentation module can be used for segmenting the generated samples if necessary.

\subsection{Generative capabilities}
\label{sec:generativeCapabilities}
The proposed method has an inherent capability to form new shapes by part feature exchange and by combining different part features. In addition, it allows integration of generative models to generate completely new parts and shapes. For this purpose, we created two variants using two different generative models: latent-space GAN (l-GAN) and VAE. l-GAN model and VAE sampling layers were integrated in between the part feature extractor and the global feature extractor to expand the system to have part generation ability -in addition to its ability to generate the global shape-. 

Latent-space GAN (l-GAN) \cite{achlioptas2017latent_pc} works in latent space instead of the actual data space. A naive GAN is placed between the Encoder and Decoder that takes part features of the dataset as real input and aims to generate fake part features that result in realistic shapes when decoded. A WGAN has also been implemented to work in the latent space (l-WGAN) to observe the differences. Gradient penalty has been applied and Discriminator has been trained more for more stable training \cite{gulrajani2017improved}.
 
While there are different AE implementations for point clouds based on PointNet, VAE based ones may fail because of the imbalance between regularization and reconstruction quality. Such models suffer from poor reconstruction/poor generation capabilities \cite{achlioptas2017latent_pc}. To overcome the imbalance problem, an additional coefficient \(\beta\) is used to weigh the regularization term. The objective function of VAE can be defined using a variational lower bound as \cite{higgins2017beta}:
\begin{equation}
\mathcal{L} = \mathbb{E}_{q_\phi (z|x)}[logp_\theta(x|z)] - \beta D_{KL}(q_\phi(z|x)||p(z))
\end{equation}

where $q$ and $p$ are data projection and generation modules with parameters $\phi$ and $\theta$ respectively and $D_{KL}$ is Kullback–Leibler divergence \cite{kullback1951}.
\section{Experimental Evaluation}
\label{sec:experimentalEvaluation}
\paragraph{Dataset:}
We used re-organized annotated ShapeNetPart dataset \cite{Yi16}, which is a subset of the highly popular ShapeNet 3D dataset \cite{shapenet2015}. It contains part labels for more than 16000 models in 16 categories and the number of parts for each category varies from 2 to 6. Each point in the point cloud sample has a semantic part label. From these 16 categories, chair, table and plane categories have been used for the study since they have the highest number of samples (3758, 5266 and 2690 samples, respectively). Each sample has a different number of points varying from 500 to 3000 points. For all the experiments, 2048 points per sample have been used, unless otherwise stated. To set all the samples the same size, random down-sampling or zero-padding have been applied. Parts can have any number of points for each model. Official train, validation and test subsets are used with 70\%, 10\% and 20\% ratios respectively. PyTorch has been used for implementation and PyTorch3D has been used for 3D operations \cite{ravi2020pytorch3d}. The training took a few hours on a NVIDIA RTX2070 GPU for the base model. Code is publicly available at \newline\url{https://github.com/cihanongun/LPMNet}
\paragraph{Distance metrics:}
Chamfer distance (CD) and Earth Mover's Distance (EMD) are the most commonly used metrics to measure the similarity of point clouds and compute the reconstruction error \cite{fan2017point}. Both these metrics are permutation invariant and work on unordered sets. Chamfer Distance is a nearest neighbor distance metric for point sets. It is the squared distance of a point in the first set to the nearest neighbor point in the second set. Chamfer Distance between two point clouds $S_1$ and $S_2$ is defined as:
\begin{equation}
\begin{split}
d_{CD}(S_1,S_2) = &\sum_{p_1 \in S_1} \underset{p_2 \in S_2}{min} \|p_1-p_2\|_2^2 + \\ &\sum_{p_2 \in S_2} \underset{p_1 \in S_1}{min} \|p_1-p_2\|_2^2
\end{split}
\end{equation}
Earth Mover's Distance (EMD) \cite{rubner2000earth} (a.k.a. Wasserstein Metric) is an algorithm to measure the effort to transport one set to another. EMD for two equal-sized point clouds $S_1$ and $S_2$ is defined as:
\begin{equation}
d_{EMD}(S_1,S_2) = \underset{\phi: S_1 \rightarrow S_2}{min}  \sum_{p \in S_1} \|p-\phi (p)\|_2
\end{equation}
where \( \phi \) is a bijection. While in practice, the exact computation of EMD is prohibitively expensive, an approximate method with reported approximation error around 1\% has been used \cite{fan2017point}.

\paragraph{The Base model:} 
The AE architecture is inspired from \cite{achlioptas2017latent_pc}. The feature extractor is a PointNet model consisting of a 3-layer MLP (64, 128, $l$) with weight sharing. Each layer is followed by a ReLU activation function and a batch normalization layer. Input and feature transform subnetworks are omitted since the samples are already aligned. It has been observed that the original 5-layer architecture has no advantage over the proposed model even with more features. The segmentation module follows a similar architecture (64, 32, 16, $k$) with weight sharing and a softmax function at the end and it is trained with a classification loss. A 3-layer architecture gives similar performance with less overfitting but the performance drops with increasing feature size. Higher number of layers cause overfitting as the data is not complex and the proposed model is trained with single class. However, a more sophisticated architecture can be employed for more complex input data. The decoder generates the point cloud model with 3 fully connected layers (1024, 2048, $n\times3$) and the first two layers are followed by a ReLU function. Fewer number of layers fail to generate high quality samples while models with higher number of layers tend to overfit to training data. A model with deconvolutional layers is also a viable option. A 5-layer (512, 256, 256, 128, 3) deconvolutional architecture has similar performance to the base model with less overfitting. However, deconvolutional model is sensitive to feature size and it fails when feature size is high (e.g. 1024). For the base model, the feature size $l$ is 128 and number of points $n$ is 2048. The system has been trained using Chamfer distance as reconstruction loss and cross-entropy loss as segmentation loss. Adam optimizer \cite{kingma2014adam} has been used with a learning rate of $5\times10^{-4}$ for 1000 epochs.

\paragraph{Experiment design:} 
To evaluate the proposed method, we have conducted a number of experiments similar to those in the literature and introduced new ones. Unless otherwise stated, the base model has been used in all experiments. Evaluation of the reconstruction performance is provided in Section \ref{sec:reconstruction}, followed by the evaluation of new model generation performance in Section \ref{sec:newModelGeneration}. The study is compared with the related works in Section \ref{sec:comparisons}. The proposed method has been tested with different input sizes to prove its robustness against low-resolution data and missing points and the results are provided in Section \ref{sec:robustness}.

\subsection{Evaluation of Reconstruction}
\label{sec:reconstruction}
\begin{figure}
\centering
\includegraphics[scale=0.30]{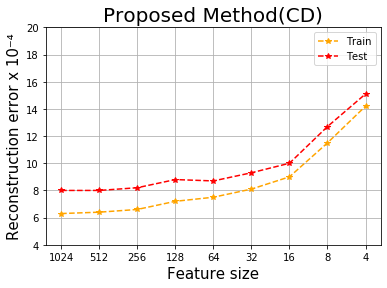}
\includegraphics[scale=0.30]{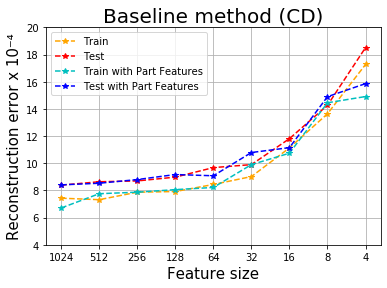}
\includegraphics[scale=0.30]{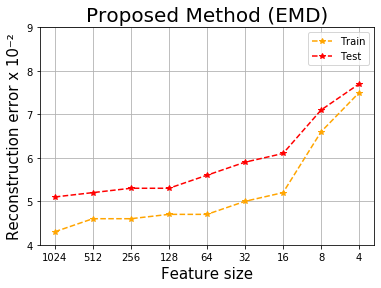}
\includegraphics[scale=0.30]{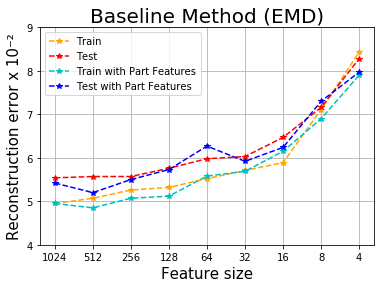}
\caption{The reconstruction losses for different feature sizes.}
\label{fig:losses}
\end{figure}

We first evaluated the effect of different feature (bottleneck) sizes. Fig. \ref{fig:losses} shows the reconstruction losses calculated using Chamfer and EMD for different feature sizes for the chair category. The proposed method and the baseline method \cite{achlioptas2017latent_pc} exhibit a similar trend that both suffer from higher reconstruction loss when the feature size is less than 128. In addition, to evaluate the effect of the part feature extractor on the reconstruction quality, the proposed part feature extractor has been integrated into the baseline method \cite{achlioptas2017latent_pc}. The results show no significant difference, supporting our claim that the global feature is not affected by the part feature extraction step. According to Fig. \ref{fig:losses}, a feature size of 128 provides a good balance to run the system with a smaller feature space without sacrificing reconstruction performance; so the feature size is set to 128 for all experiments. 

\begin{figure}[h]
\centering
\includegraphics[width = 0.48\textwidth]{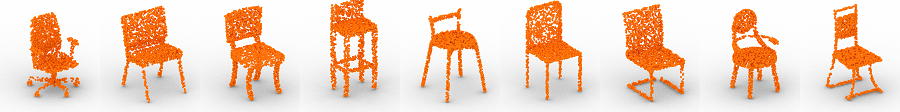}\\
\includegraphics[width = 0.48\textwidth]{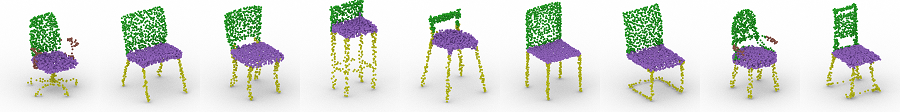}\\
\includegraphics[width = 0.48\textwidth]{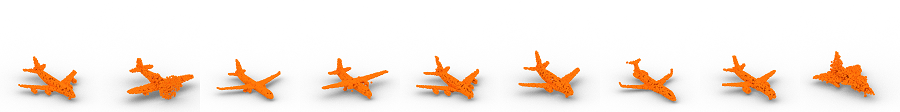}\\
\includegraphics[width = 0.48\textwidth]{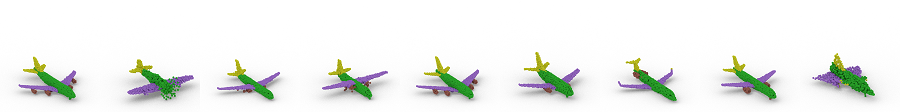}\\
\includegraphics[width = 0.48\textwidth]{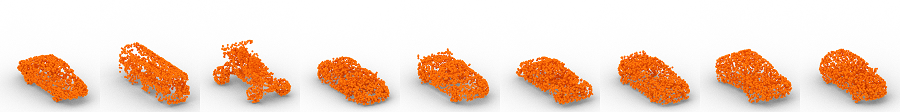}\\
\includegraphics[width = 0.48\textwidth]{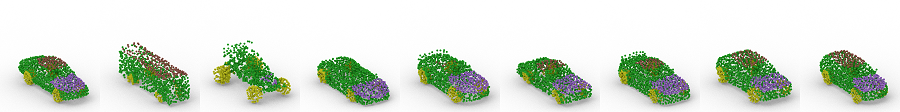}\\
\includegraphics[width = 0.48\textwidth]{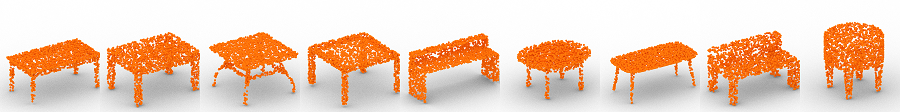}\\
\includegraphics[width = 0.48\textwidth]{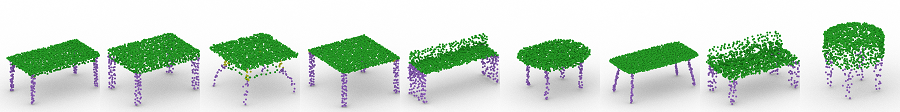}\\
\caption{The reconstruction results of the proposed model. For each object class, the first row shows the samples from the unlabeled test set and the second row shows the corresponding reconstructions.}
\label{fig:rec}
\end{figure}

%The first set of experiments aim to evaluate the reconstruction capabilities of the proposed AE architecture with different global and part feature vector sizes. 
The reconstruction results on the test set can be seen in Fig. \ref{fig:rec}. Visual results indicate good reconstruction performance with minor loss. 

Part interpolation and part exchange experiments aim to validate that a regularized part feature space can extract the part features separately and parts can be exchanged between different generated shapes. Then, we show that different parts from different shapes can be used to compose new shapes.

\begin{figure*}
\centering
\rotatebox{90}{\hspace{0.5cm} Global}
\includegraphics[width = 0.9\textwidth]{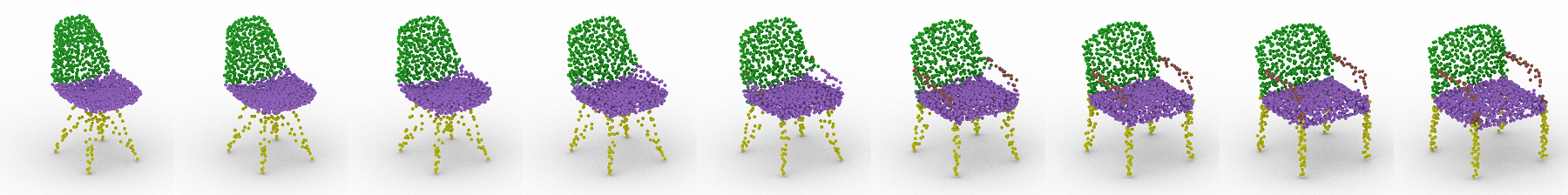}\\
\rotatebox{90}{\hspace{0.5cm} Back}
\includegraphics[width = 0.9\textwidth]{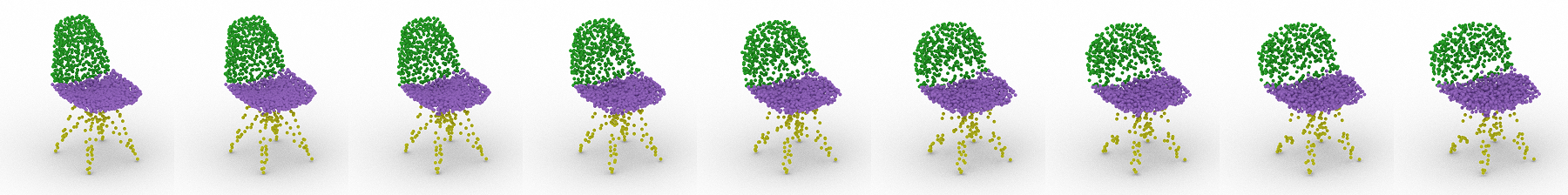}\\
\rotatebox{90}{\hspace{0.5cm} Seat}
\includegraphics[width = 0.9\textwidth]{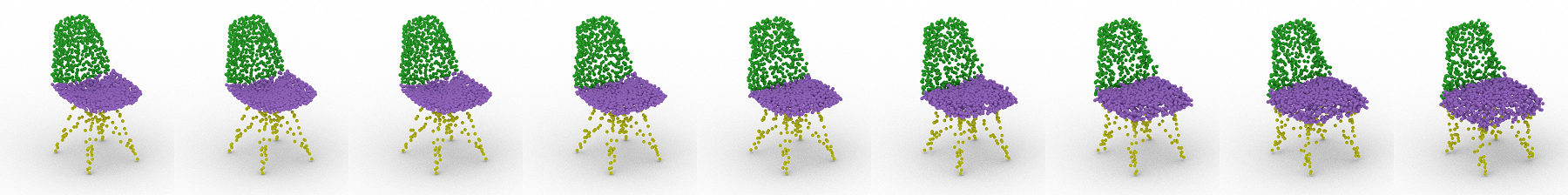}\\
\rotatebox{90}{\hspace{0.5cm} Leg}
\includegraphics[width = 0.9\textwidth]{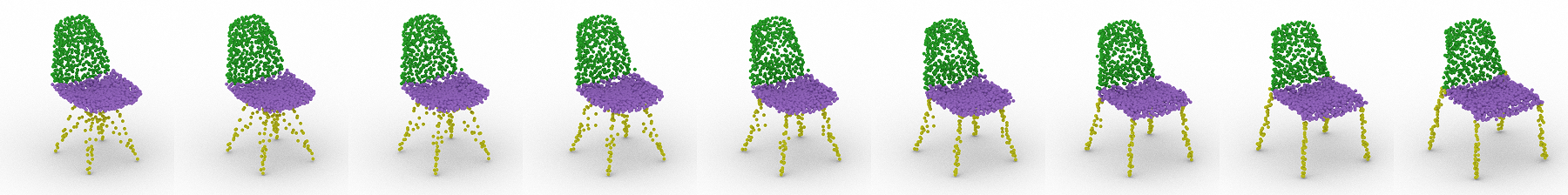}\\
\rotatebox{90}{\hspace{0.5cm} Arm}
\includegraphics[width = 0.9\textwidth]{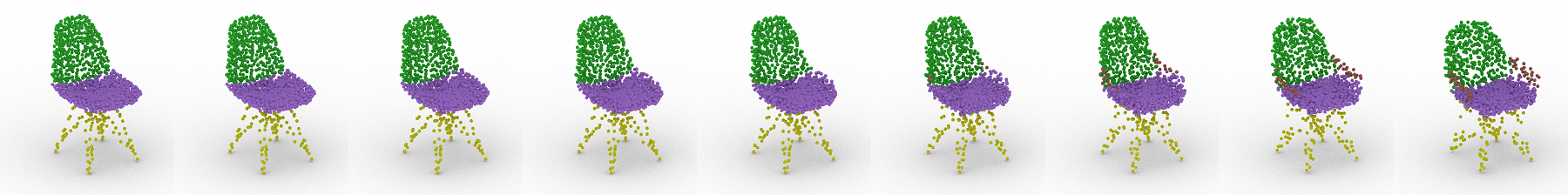}\\
\caption{Part interpolation between two shapes. The first row is global shape interpolation between two shapes (leftmost and rightmost). Other rows are single part interpolations where only the corresponding part feature is interpolated while features of other parts are kept the same.}
\label{fig:interpmerge}
\end{figure*}

\paragraph{Part interpolation and part exchange:}
By modifying the part feature, shape of a respective part could be changed in isolation, keeping the other parts the same. To prove this claim, we apply part interpolations for all parts separately and show the results in Fig. \ref{fig:interpmerge}. Global feature interpolation results in a smooth interpolation between two different shapes reflecting a regular and continuous latent space. Part feature interpolation interpolates only a specific part and assembles the new part into the existing sample. Here it can be seen that it is not a naive part assembly transplanting a part into another shape. Latent space represents the semantic properties of a part so it generates a part that matches better to the new shape by preserving semantic properties. For example, using the leg part feature of a four-legged chair with an office chair having wheels generates the same office chair with four legs instead of wheels. However, the leg part will not be the same as the source chair since it would not be a good fit for the target office chair. The office chair is now generated with four legs which are in better harmony with the rest of the shape resulting in a more realistic looking chair. Results for other classes can be seen in Fig. \ref{fig:apinterp}.

\begin{figure}[h]
\begin{tabular}{c @{} c @{} c @{} c @{} c}
Back & Seat & Leg & Arm & \textbf{Composition}\\ 
\includegraphics[width = 0.09\textwidth]{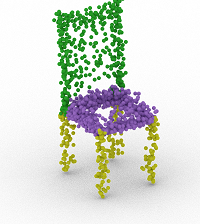}&
\includegraphics[width = 0.09\textwidth]{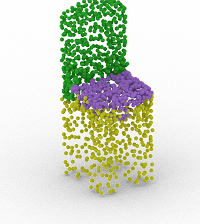}&
\includegraphics[width = 0.09\textwidth]{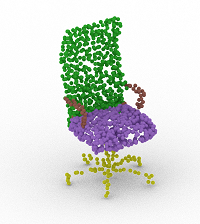}&
\includegraphics[width = 0.09\textwidth]{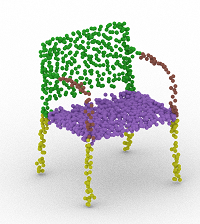}&
\includegraphics[width = 0.09\textwidth]{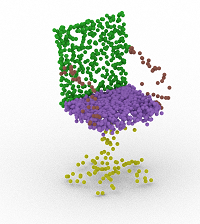}\\
Body & Wings & Tail & Engines & \textbf{Composition} \\
\includegraphics[width = 0.09\textwidth, trim=0 0 0 2cm, clip]{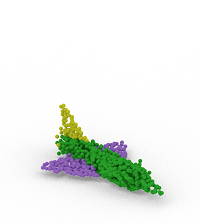}&
\includegraphics[width = 0.09\textwidth, trim=0 0 0 2cm, clip]{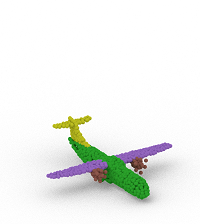}&
\includegraphics[width = 0.09\textwidth, trim=0 0 0 2cm, clip]{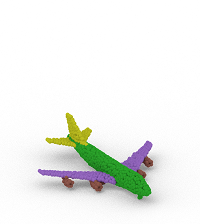}&
\includegraphics[width = 0.09\textwidth, trim=0 0 0 2cm, clip]{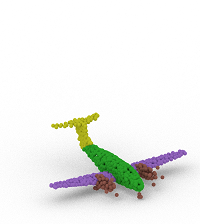}&
\includegraphics[width = 0.09\textwidth, trim=0 0 0 2cm, clip]{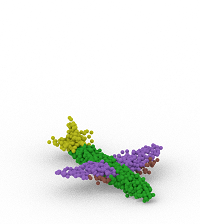}\\
\multicolumn{2}{c}{Top}  & \multicolumn{2}{c}{Foot} & \textbf{Composition}\\
\multicolumn{2}{c}{\includegraphics[width = 0.09\textwidth, trim=0 0 0 1cm, clip]{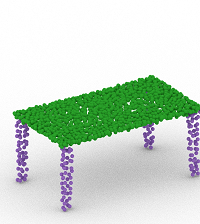}} &
\multicolumn{2}{c}{\includegraphics[width = 0.09\textwidth, trim=0 0 0 1cm, clip]{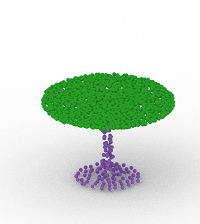}}&
\includegraphics[width = 0.09\textwidth, trim=0 0 0 1cm, clip]{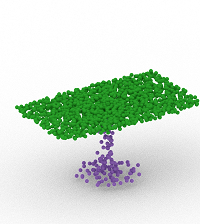}
\end{tabular}
\caption{Part features from different samples are combined together to form a new shape. Parts may not be exactly the same as they are in source shapes for a more coherent composition.}
\label{fig:compo}
\end{figure}

\paragraph{Composition of separate parts:}
%In the proposed architecture, the parts are expected to be independent of the global shape.
In the proposed architecture, part features can be extracted independently for composing new shapes. Parts are expected to be independent of each other to form new global shapes. To test the validity of the independence assumption of the parts, different part features from different models are merged to obtain a global feature. This global feature is then used to generate a global shape with these parts. Part features carry the semantics of corresponding parts. With global feature extraction, a global feature is formed from part features that gathers all semantics together. The decoder generates a global shape from global feature that represents all semantics. Sample results can be seen in Fig. \ref{fig:compo}. A new shape is formed by the selected parts without any need for assembling the parts together with affine transformations. It has to be noted that the parts may not be exactly the same as they are in source shapes. The parts may get modified for a more coherent composition. The experiments validate that new samples can be generated using different parts from different shapes.

\subsection{Evaluation of New Model Generation}
\label{sec:newModelGeneration}
The method can be extended to have generative capabilities by integration of generative models. In this section, we evaluate the generation of new global shapes and parts by integrating two separate models: GAN and Variational Autoencoder (VAE). 

Latent-space GAN based architecture \cite{achlioptas2017latent_pc} uses encoded data as its input and output. Generator is a 3-layer MLP (128, $l$, $k\times l$) for $k$ parts and the Discriminator mirrors the Generator. Generator input is a 128-dimensional vector sampled from a Normal distribution. l-GAN has been trained using Adam optimizer with a first-moment value of 0.5 and learning rates of $5\times10^{-4}$ and $1\times10^{-4}$ for Generator and Discriminator respectively. GAN has been trained with the pretrained model to extract and decode features. WGAN follows the same architecture with a different objective function.

\begin{figure*}
\rotatebox{90}{\hspace{0.5cm} VAE}
\includegraphics[width=0.95\textwidth]{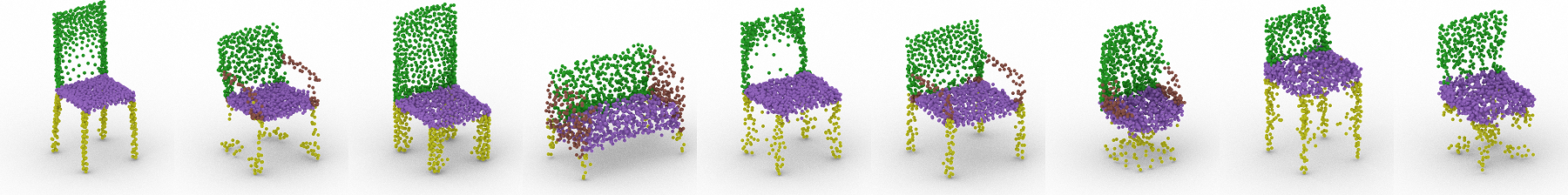}\\
\rotatebox{90}{\hspace{0.5cm} GAN}
\includegraphics[width=0.95\textwidth]{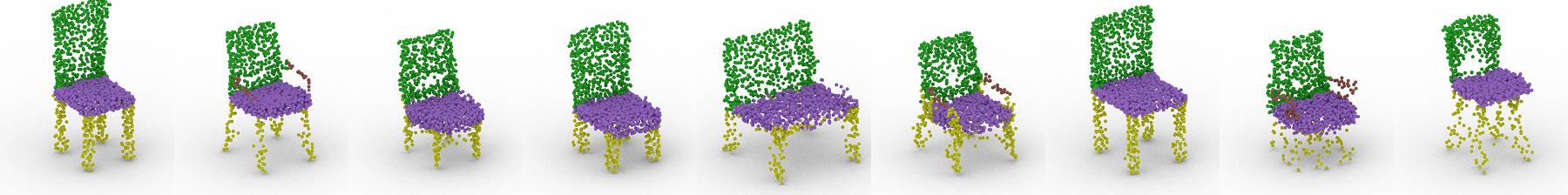}\\
\rotatebox{90}{\hspace{0.5cm} WGAN}
\includegraphics[width=0.95\textwidth]{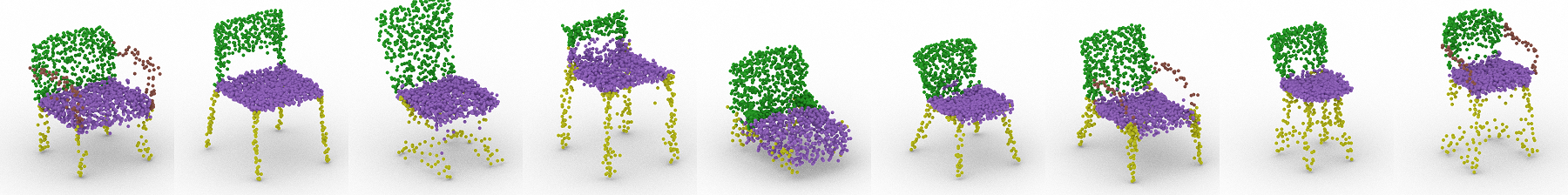}
\caption{Samples from generative models. VAE provides good reconstruction and generation capabilities. While standard GAN is able to generate good results, it suffers from lack of diversity. WGAN generates more diverse results.}
\label{fig:gens}
\end{figure*}

VAE based architecture follows the base model with an exception of the sampling layers, which are now fully connected layers to generate mean and sigma values.  Regularization term has been normalized with input dimension and \(\beta\) parameter has been set to $0.1$ since it provides a good balance between reconstruction and generation quality. Reparametrization trick has been employed and the system has been trained using Adam optimizer \cite{kingma2014adam} with a learning rate of $10^{-3}$ for 10000 epochs. For new data generation, latent codes have been sampled from a Normal distribution. Generated samples can be seen in Fig. \ref{fig:gens} for chair class and Fig. \ref{fig:apgen} for plane, car and table classes.

For the evaluation of generative models, we have used the following metrics: Coverage (Cov), Minimum Matching Distance (MMD) and Jensen–Shannon Divergence (JSD) \cite{achlioptas2017latent_pc}. Cov measures the representation of a point cloud set $S_2$ in set $S_1$. It is the fraction of point clouds in one set that is matched to others by finding the nearest neighbor. MMD is the average of distances between the matched point clouds in different sets. JSD is the distance between 2 probability distributions, it is derived from  Kullback–Leibler divergence \cite{kullback1951}. In this scope, it is used as a measure of occupation of similar locations in 3D coordinate space between two point cloud sets. MMD and Cov have been calculated using both CD and EMD. Total Mutual Difference (TMD) \cite{wu_2020_ECCV} is used to measure the diversity of the generated shapes when one or more parts are changed. It is calculated by finding the average Chamfer distance of all shapes with generated parts for a given input shape. A higher score is better for Coverage and TMD and a lower score is better for MMD and JSD.

New samples are generated by five different approaches: (\textit{i}) \textit{part feature exchange}: randomly exchanging part features between different samples, (\text{ii}) \textit{part feature composition}: composing new shapes by combining different part features from different random samples, (\textit{iii}) \textit{VAE}: new shapes are generated by sampling from a Normal distribution using VAE, (\textit{iv}) \textit{GAN}: GAN is used after training to randomly generate new shapes, (\textit{v}) \textit{WGAN}: WGAN is used instead of GAN for more diversity and more stable training. All models have been trained with CD and EMD. A sample set is formed by generation results, which is 3 times the size of the test set. Results can be seen in Table \ref{table:gens}. As expected, the results are in favor of the models trained with the same distance metric as the evaluation method. Part exchange has the lowest distance score with a high coverage. This is expected since only a single part per sample is different from the reference test set. Also, high coverage supports the similarity between the test set and the part-exchange set. The random part composition approach exhibits good diversity and novelty comparable with the generative models. GAN implementation exhibits overfitting and collapses to a single mode especially when trained with EMD distance. WGAN achieves better diversity as expected with better coverage scores than GAN. VAE performs similar to WGAN indicating good sampling capability besides reconstruction. \textit{Plane} class has lower MMD and JSD distance scores than other classes since the plane models are smaller, more dense, less diverse and occupy less area. The results show that different alternatives are successful at different aspects and they may serve different tasks better depending on the quality, diversity or complexity requirements of a particular task.

\begin{table*}
\small
\begin{center}
\caption{Evaluation of generative models based on Minimum Matching Distance (MMD), Coverage (Cov), and Jensen-Shannon Divergence (JSD$\times10^{-2}$). Both CD $(\times10^{-4})$ and EMD $(\times10^{-2})$ metrics are used for evaluation. CN is the part-assembly based approach CompoNet\cite{Schor_2019_ICCV}. Ach. is the best generative method (l-WGAN) reported in the baseline study Achlioptas et al.\cite{achlioptas2017latent_pc}. Tree-GAN results are reported in \cite{treegan}. The best results, among only the generative models, are marked in bold.}
\label{table:gens}
\begin{tabular}{l c c c c c | c c c c c | c c c c c}
 &\multicolumn{5}{c}{\textit{chair}} &\multicolumn{5}{c}{\textit{table}} &\multicolumn{5}{c}{\textit{plane}}\\
 \hline
 &\multicolumn{2}{c}{MMD} &\multicolumn{2}{c}{\% Cov} & & \multicolumn{2}{c}{MMD} &\multicolumn{2}{c}{\% Cov} & & \multicolumn{2}{c}{MMD} &\multicolumn{2}{c}{\% Cov}\\ 
Model&CD&EM&CD&EMD&JSD&CD&EMD&CD&EMD&JSD&CD&EMD&CD&EMD&JSD\\
\hline
\multicolumn{16}{c}{Trained with CD} \\
\hline
Exc. & 14.39 & 9.53 & 72.65 & 32.03 & 4.88 & 13.45 & 7.69 & 70.31 & 34.37 & 3.13 & 3.90 & 5.85 & 69.53 & 14.06 & 3.73\\ 
Comp. & 17.50 & 9.86 & 56.25 & 25.78 & 5.58 & 15.77 & 7.83 & 67.19 & 32.03 & 3.81 & 4.40 & 5.99 & 60.93 & 11.71 & 4.18\\ 
\hline
VAE & \textbf{14.77} & 10.24 & \textbf{69.53} & 28.12 & 6.74 & \textbf{13.62} & 7.96 & \textbf{71.87} & \textbf{40.62} & \textbf{3.40} & \textbf{3.43} & 6.41 & 59.59 & 14.84 & 5.64\\
GAN & 22.41 & 10.39 & 34.37 & 19.53 & 8.97 & 33.38 & 9.94 & 21.09 & 14.84 & 8.00 & 6.39 & 6.31 & 24.21 & 7.81 & 5.98\\ 
WGAN & 15.76 & \textbf{9.64} & 52.34 & 21.87 & \textbf{5.88} & 16.40 & \textbf{7.95} & 60.15 & 35.16 & 4.93 & 4.76 & \textbf{5.76} & 60.93 & 15.62 & \textbf{4.17}\\ 
CompoNet \cite{Schor_2019_ICCV} & 40.63 & 10.11 & 28.90 & \textbf{32.03} & 7.65 & 87.07 & 14.14 & 30.46 & 14.85 & 22.99 & 20.02 & 8.41 & 19.53 & 16.4 & 17.83\\
Tree-GAN \cite{treegan} & 16.00 & 10.10 & 58.00 & 30.00 & 11.90 & 18.00 & 10.70 & 66.00 & 39.00 & 10.05 & 4.00 & 6.80 & \textbf{61.00} & \textbf{20.00} & 9.70\\
\hline
\multicolumn{16}{c}{Trained with EMD} \\
\hline
Exc. & 18.10 & 6.64 & 71.09 & 76.56 & 1.66 & 17.01 & 5.94 & 75.00 & 78.12 & 1.99 & 4.45 & 3.80 & 72.65 & 67.18 & 2.05\\ 
Comp. & 22.14 & 7.32 & 56.25 & 61.71 & 2.02 & 19.41 & 6.48 & 70.31 & 72.65 & 2.46 & 5.41 & 4.21 & 59.37 & 53.12 & 2.74\\ 
\hline
VAE & 23.87 & 7.84 & 55.47 & \textbf{67.19} & 4.28 & 23.58 & 7.23 & 50.78 & 60.15 & 4.32 & \textbf{5.29} & \textbf{4.14} & 57.04 & 53.12 & 3.54\\
GAN & 34.48 & 8.99 & 23.43 & 24.21 & 6.41 & 32.87 & 8.34 & 31.25 & 38.28 & 6.10 & 6.23 & 4.61 & 42.96 & 35.15 & 3.51\\ 
WGAN & 23.11 & 7.44 & 56.25 & 60.93 & 3.01 & \textbf{20.71} & 6.79 & 66.40 & 71.87 & 3.32 & 6.03 & 4.31 & \textbf{57.07} & 52.34 & \textbf{2.62}\\
Ach. et al. \cite{achlioptas2017latent_pc} & \textbf{21.95} & \textbf{7.06} & \textbf{70.31} & 66.4 & \textbf{2.74} & 20.75 & \textbf{6.64} & \textbf{69.53} & \textbf{73.43} & \textbf{2.76} & 6.49 & 4.21 & 57.03 & \textbf{60.93} & 3.25\\
\hline
\end{tabular}
\end{center}
\end{table*}

\begin{figure}[h]
\centering
\rotatebox{90}{\hspace{0.3cm} Back}
\includegraphics[width = 0.45\textwidth]{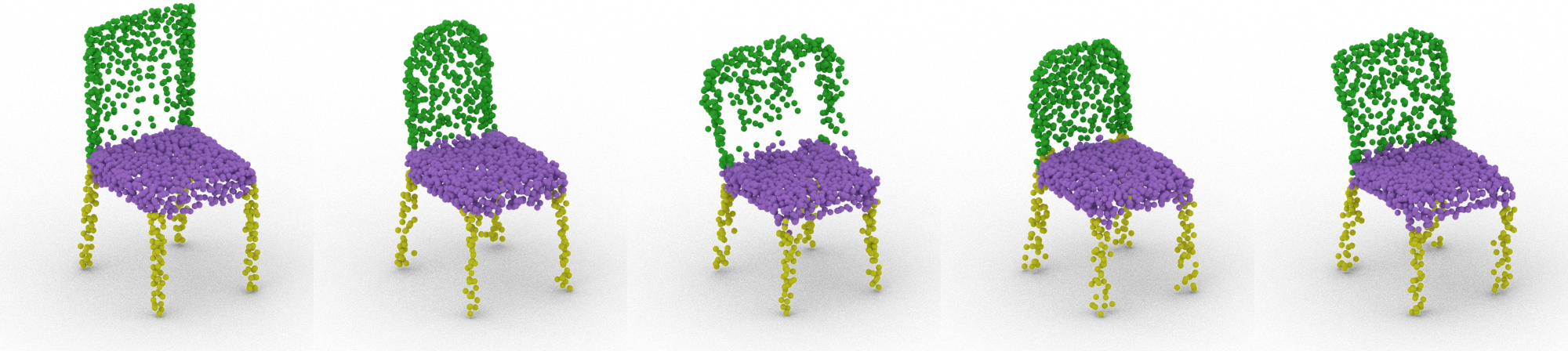}\\
\rotatebox{90}{\hspace{0.3cm} Seat}
\includegraphics[width = 0.45\textwidth]{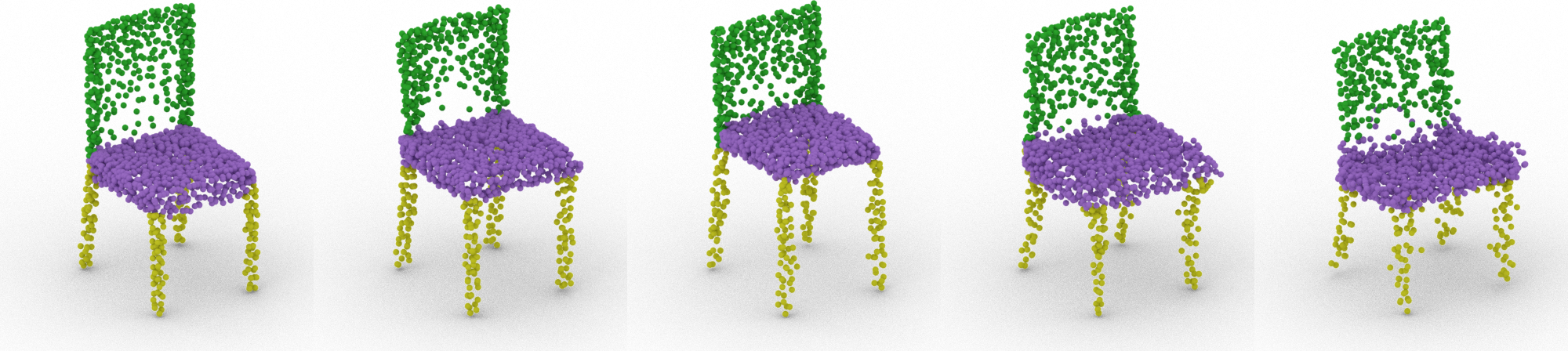}\\
\rotatebox{90}{\hspace{0.4cm} Leg}
\includegraphics[width = 0.45\textwidth]{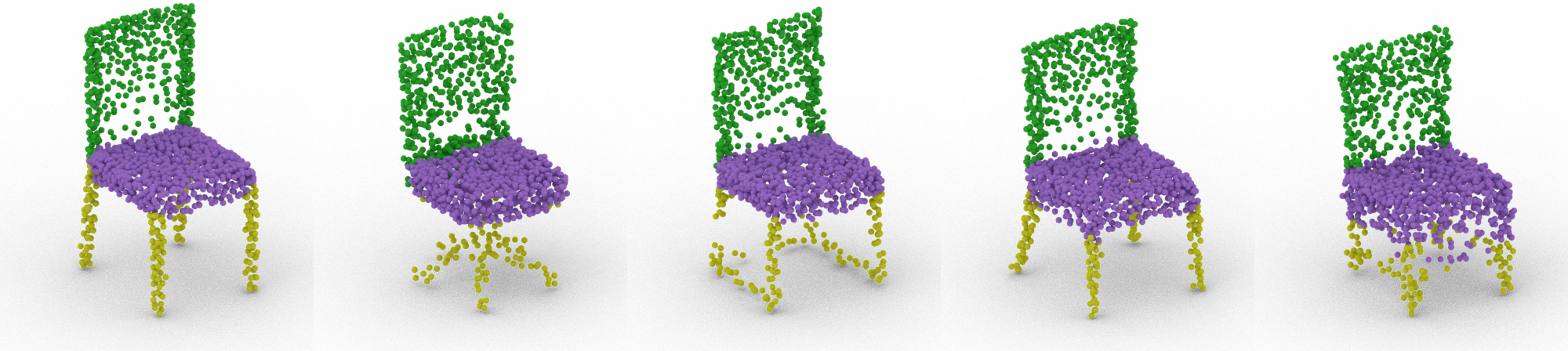}\\
\rotatebox{90}{\hspace{0.4cm} Arm}
\includegraphics[width = 0.45\textwidth]{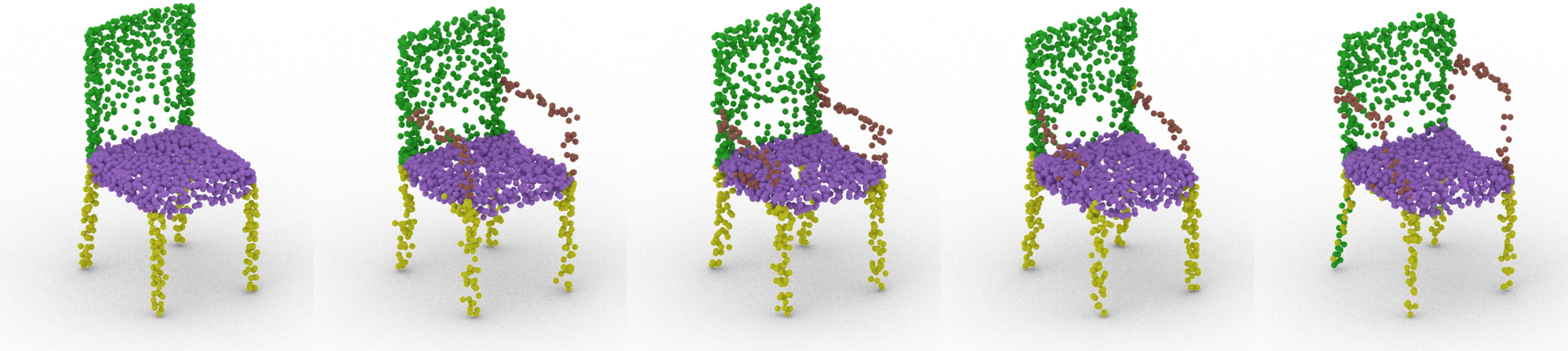}
\caption{Samples from part exchange and generation for an existing model (most left).}
\label{fig:exc}
\end{figure}

TMD is calculated by generating 10 samples for each shape by changing one or more parts while keeping the other parts the same. TMD  results for the chair class are reported in Table \ref{table:tmd}, and sample visualizations are provided in Fig. \ref{fig:exc}. As expected, for all models, TMD score gets higher when higher number of parts are generated. The exchange approach performs the best since it exchanges the parts with the already existing ones in the dataset. Other methods generate new parts from scratch, thus showing less diversity. The results for table and plane classes are provided in Table \ref{table:tmd_table} and \ref{table:tmd_plane} respectively.

\begin{table}
\begin{center}
\caption{Total Mutual Difference (TMD$\times10^{-2}$) \cite{wu_2020_ECCV} scores for part exchange and generation. One or more parts are changed by keeping the others the same.}
\label{table:tmd}
\begin{tabular}{l c c c c}
\hline
 & \multicolumn{4}{c}{\# of changing parts} \\ 
Model & 1 & 2 & 3 & 4\\
\hline
Exchange & 1.31 & 3.47 & 4.66 & 4.85 \\ 
VAE & 1.06 & 2.54 & 3.33 & \textbf{3.54}  \\
l-GAN & 0.79 & 1.96 & 2.41 & 2.60 \\ 
l-WGAN & 1.22 & 2.53 & \textbf{3.38} & 3.48 \\ 
\hline
\hline
Wu et al. \cite{wu_2020_ECCV}& \textbf{2.28} & \textbf{2.81} & 2.96 & 3.19 \\
\hline
\end{tabular}
\end{center}
\end{table}

\subsection{Comparison with related works}
\label{sec:comparisons}

\begin{figure*}[h]
\rotatebox{90}{\hspace{0.2cm} Ach. et al.}
\includegraphics[width=0.95\textwidth]{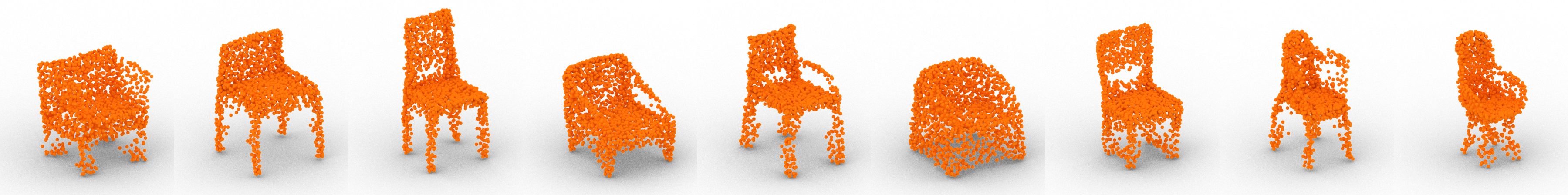}\\
\rotatebox{90}{\hspace{0.2cm} CompoNet}
\includegraphics[width=0.95\textwidth]{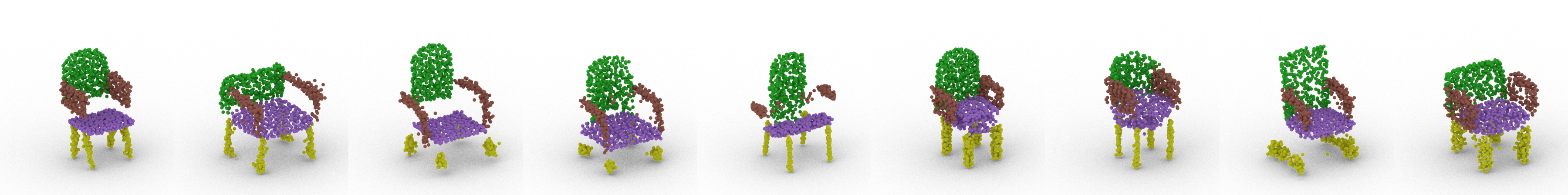}\\
\rotatebox{90}{\hspace{0.1cm} StructureNet}
\includegraphics[width=0.95\textwidth]{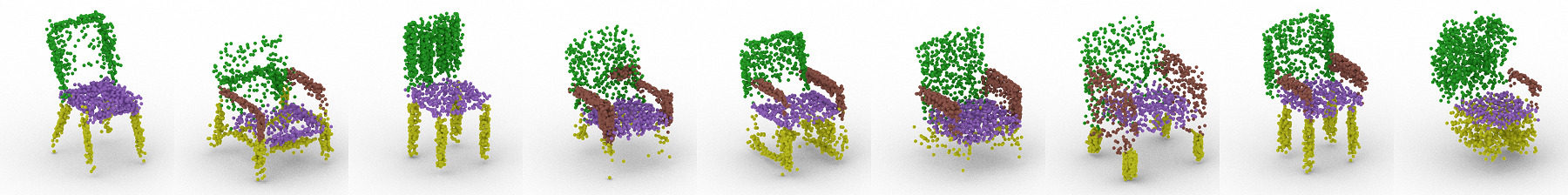}\\
\rotatebox{90}{\hspace{0.5cm} Ours}
\includegraphics[width=0.95\textwidth]{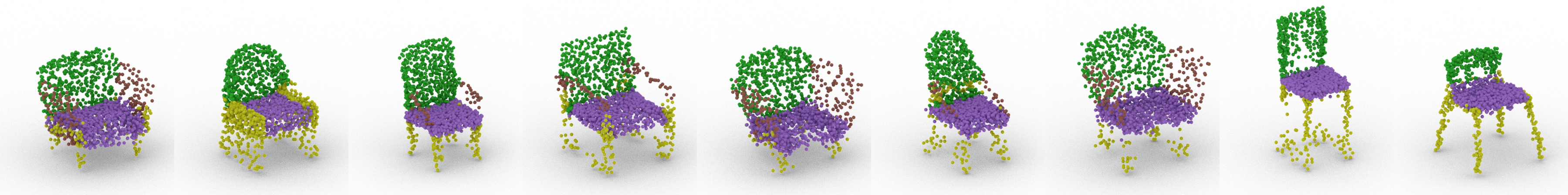}
\caption{Randomly generated samples by different methods; Achlioptas et al. \cite{achlioptas2017latent_pc}, CompoNet \cite{Schor_2019_ICCV}, StructureNet \cite{Mo2019StructureNetHG} and our model. Achlioptas et al. considers only the global shape, CompoNet has difficulty assembling and connecting the generated parts. StructureNet can generate more diverse structures but suffers from structural noise causing implausible structures.}
\label{fig:benchs}
\end{figure*}

The results of the proposed work and related works are provided in Table \ref{table:gens}. CompoNet \cite{Schor_2019_ICCV} is a part-assembly based approach. It has separate Autoencoders trained with CD for different parts and these individually generated parts are then brought together by a part-assembly network. The results show that, the proposed method outperforms CompoNet in all cases. Although part generation of this method is satisfactory, the part-assembly step generates incoherent global shapes, which fail to exhibit seamless connection between parts. Also, points are not distributed evenly across the global shape as there are fixed number of points per part. The best generative model in the baseline study (l-WGAN trained with EMD) is selected for the comparison \cite{achlioptas2017latent_pc}. As expected, the proposed method has similar performance with the baseline method, since both these methods become equivalent for global shape generation. However, the proposed method has additional part-based capabilities as mentioned above. Tree-GAN \cite{treegan} has comparable results with the other generative models. However, it cannot be evaluated with regards to part exchange and composition performance as it lacks reconstruction abilities. Its MMD and Coverage results are inferior for \textit{chair} and \textit{table} classes. While it has better results for Coverage of \textit{plane} class, the difference is only marginal. StructureNet uses a fine-grained, hierarchical dataset for structure encoding, hence its results cannot be evaluated on the dataset used in these experiments. To allow comparisons with StructureNet, we conducted a separate experiment, by training our method on their dataset, and presented the results in Section \ref{sec:apcomp} as supplementary comparisons.

The qualitative results can be seen in Fig. \ref{fig:benchs}. The proposed method and the baseline (Achlioptas et al. \cite{achlioptas2017latent_pc}) show similar generation quality and diversity. However, the baseline method does not have any part information and only considers global shapes. The part-assembly based CompoNet \cite{Schor_2019_ICCV} is able to generate parts separately, but it has difficulty assembling and connecting the generated parts. By using a part-based holistic approach, (\textit{i}) the proposed method can handle separate parts, which is a capability lacking in \cite{achlioptas2017latent_pc} and (\textit{ii}) it also generates a complete coherent global shape in unison while handling separate parts which is different to the two stage approach in \cite{Schor_2019_ICCV}. This eliminates the need for a separate part-assembly network and potential problems associated with part-assembly. StructureNet \cite{Mo2019StructureNetHG} (results are downsampled to the same number of points for fair comparison) generates diverse structures including asymmetric ones. However, the generated samples suffer from structural noise causing implausible shapes. Also, representing all parts with the same number of points leads to better quality for small parts than large parts, especially becoming evident in low resolutions. 

For the evaluation of shape completion capability, Total Mutual Difference (TMD) results are reported in Table \ref{table:tmd} by regenerating one or more parts. Wu et al. \cite{wu_2020_ECCV} is a shape completion network  which completes the partial shapes with missing parts by generating multiple outputs. The proposed method has lower scores for few missing parts, but exhibits higher scores when there are higher number of missing part. However, it has to be noted that TMD evaluates the diversity of the whole shape and not only the generated part. While completing a shape with a new part, the method in \cite{wu_2020_ECCV} also causes changes in the other parts of the shape, which results in an increase in TMD score. This observation is supported by the low TMD score variance with respect to the different number of missing parts for this method. 

\subsection{Robustness Against Different Input Sizes}
\label{sec:robustness}
The same shape can be defined by using different number of points. So, the method is expected to have the ability to process different input point cloud sizes (resolutions) and give similar outputs. In this section, we evaluate the performance of the proposed method against different input sizes and compare the critical points extracted from different input sizes.

To define a global feature, a feature extractor first detects the critical points, which are the most important points in a point cloud sample. The critical point set is the minimum number of points defining the shape. For example, the corner points are the critical points that define a triangle. The feature set defines the semantics of the shape irrespective of the resolution, so a higher resolution sample also results in the same global feature set (i.e., the corners of a triangle). 

\begin{figure*}
\begin{tabular}{|c|c|}
\hline
\includegraphics[width = 0.47\textwidth]{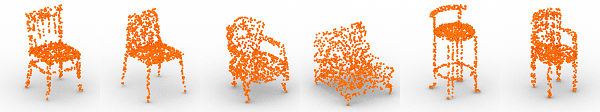} &
\includegraphics[width = 0.47\textwidth]{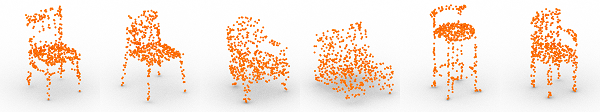}\\
\includegraphics[width = 0.47\textwidth]{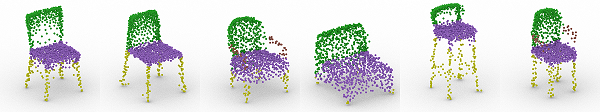} &
\includegraphics[width = 0.47\textwidth]{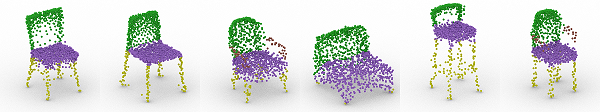}\\
\hline
\includegraphics[width = 0.47\textwidth]{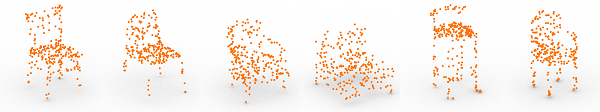} &
\includegraphics[width = 0.47\textwidth]{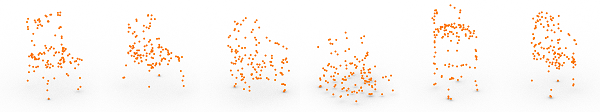}\\
\includegraphics[width = 0.47\textwidth]{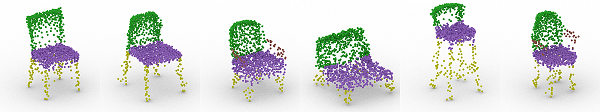} &
\includegraphics[width = 0.47\textwidth]{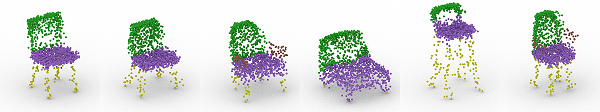}\\
\hline
\end{tabular}
\caption{Reconstruction results from 1024 (top-left), 512 (top-right), 256 (bottom-left) and 128 (bottom-right) points to 2048 points.}
\label{fig:upsample}
\end{figure*}

The proposed method is expected to extract the same feature set for a shape defined with different number of points. These features can then be decoded to reconstruct the shape at any size. To test this, the original input has been randomly downsampled to 1024, 512, 256 and 128 points from 2048 points. Then these samples have been zero-padded to obtain 2048 points and the zero-padded points have been labeled as part 0. Then, these samples have been fed into the pretrained network to reconstruct the shape. Since the network ignores part 0 for feature extraction, it extracts the same features for all input dimensions. The results in Fig. \ref{fig:upsample} shows that the system can handle different input dimensions by giving the same features for the same shapes. The results are not affected by the lack of zero-padded samples during training.  Also, this approach can serve as an upsampling network without training from scratch. It has to be noted that a lower number of input points result in poorer reconstructions since some critical points vanish due to random downsampling. Removing batch normalization layers improves robustness with more independent point features.

\section{Conclusions}
\label{sec:conclusions}
In this paper, a generic part-aware architecture allowing exchanging of parts between different models and generating new point cloud models and parts has been proposed. The proposed system is based on a single network and does not need separate networks for each part or an additional network to assemble them to form a new shape. The system has been proven to work with different object categories having different numbers of parts and varying sizes. The system provides an end-to-end solution for unlabeled data with the integration od a segmentation module. It has been shown that GANs and VAEs can be integrated into the proposed method to generate new parts and models.

In the proposed method, while a part feature represents the corresponding part in a global shape, the decoder takes a global feature as input and outputs a global shape. While the method cannot reconstruct the parts separately, this is not considered to be a significant limitation as the ultimate aim in most applications is to form a global shape. To reconstruct the parts separately, the method must be trained with parts separately from scratch. Then, the global shape can be constructed from the parts by a composition model similar to those in the literature.  Part modification and generation are complementary operations to get the global shapes. 

In some cases, reconstruction of uncommon samples (e.g., asymmetrical samples, samples with incorrect labels) may fail, especially if they are only encountered in the test set. These samples are considered to be outliers by the network and they have limited effect in the learning and hence they are not represented effectively by the network. Processing outliers is a common and challenging problem for neural networks based systems.

\section*{Acknowledgements}
This work has been supported by Middle East Technical University Scientific Research Projects Coordination Unit under grant number GAP-704-2020-10071.
%%Vancouver style references.
\bibliographystyle{cag-num-names}
\bibliography{refs}

\begin{thebibliography}{32}
\providecommand{\natexlab}[1]{#1}
\providecommand{\url}[1]{\texttt{#1}}
\providecommand{\href}[2]{#2}
\providecommand{\path}[1]{#1}
\providecommand{\eprint}[1]{\href{http://arxiv.org/abs/#1}{\path{#1}}}
\providecommand{\DOIprefix}{doi:}
\providecommand{\ArXivprefix}{arXiv:}
\providecommand{\URLprefix}{URL: }
\providecommand{\Pubmedprefix}{pmid:}
\providecommand{\doi}[1]{\href{http://dx.doi.org/#1}{\path{#1}}}
\providecommand{\Pubmed}[1]{\href{pmid:#1}{\path{#1}}}
\providecommand{\BIBand}{and}
\providecommand{\bibinfo}[2]{#2}
\ifx\xfnm\undefined \def\xfnm[#1]{\unskip,\space#1}\fi
%Type = Inproceedings
\bibitem[{{Charles} et~al.(2017){Charles}, {Su}, {Kaichun} and
  {Guibas}}]{qi2016pointnet}
\bibinfo{author}{{Charles}\xfnm[ RQ]}, \bibinfo{author}{{Su}\xfnm[ H]},
  \bibinfo{author}{{Kaichun}\xfnm[ M]}, \bibinfo{author}{{Guibas}\xfnm[ LJ]}.
\newblock \bibinfo{title}{Pointnet: Deep learning on point sets for 3d
  classification and segmentation}.
\newblock In: \bibinfo{booktitle}{2017 IEEE Conference on Computer Vision and
  Pattern Recognition (CVPR)}. \bibinfo{year}{2017}, p.
  \bibinfo{pages}{77--85}.
\newblock \DOIprefix\doi{10.1109/CVPR.2017.16}.
%Type = Inproceedings
\bibitem[{Qi et~al.(2017)Qi, Yi, Su and Guibas}]{qi2017pointnetplusplus}
\bibinfo{author}{Qi\xfnm[ CR]}, \bibinfo{author}{Yi\xfnm[ L]},
  \bibinfo{author}{Su\xfnm[ H]}, \bibinfo{author}{Guibas\xfnm[ LJ]}.
\newblock \bibinfo{title}{Pointnet++: Deep hierarchical feature learning on
  point sets in a metric space}.
\newblock In: \bibinfo{booktitle}{Neural Information Processing Systems}.
  \bibinfo{year}{2017},.
%Type = Inproceedings
\bibitem[{Park et~al.(2019)Park, Florence, Straub, Newcombe and
  Lovegrove}]{park2019deepsdf}
\bibinfo{author}{Park\xfnm[ JJ]}, \bibinfo{author}{Florence\xfnm[ P]},
  \bibinfo{author}{Straub\xfnm[ J]}, \bibinfo{author}{Newcombe\xfnm[ R]},
  \bibinfo{author}{Lovegrove\xfnm[ S]}.
\newblock \bibinfo{title}{Deepsdf: Learning continuous signed distance
  functions for shape representation}.
\newblock In: \bibinfo{booktitle}{Proceedings of the IEEE Conference on
  Computer Vision and Pattern Recognition}. \bibinfo{year}{2019}, p.
  \bibinfo{pages}{165--174}.
%Type = Inproceedings
\bibitem[{Wu et~al.(2019)Wu, Qi and Fuxin}]{pointconv}
\bibinfo{author}{Wu\xfnm[ W]}, \bibinfo{author}{Qi\xfnm[ Z]},
  \bibinfo{author}{Fuxin\xfnm[ L]}.
\newblock \bibinfo{title}{Pointconv: Deep convolutional networks on 3d point
  clouds}.
\newblock In: \bibinfo{booktitle}{Proceedings of the IEEE Conference on
  Computer Vision and Pattern Recognition}. \bibinfo{year}{2019}, p.
  \bibinfo{pages}{9621--9630}.
%Type = Inproceedings
\bibitem[{Thomas et~al.(2019)Thomas, Qi, Deschaud, Marcotegui, Goulette and
  Guibas}]{kpconv}
\bibinfo{author}{Thomas\xfnm[ H]}, \bibinfo{author}{Qi\xfnm[ CR]},
  \bibinfo{author}{Deschaud\xfnm[ JE]}, \bibinfo{author}{Marcotegui\xfnm[ B]},
  \bibinfo{author}{Goulette\xfnm[ F]}, \bibinfo{author}{Guibas\xfnm[ LJ]}.
\newblock \bibinfo{title}{Kpconv: Flexible and deformable convolution for point
  clouds}.
\newblock In: \bibinfo{booktitle}{Proceedings of the IEEE International
  Conference on Computer Vision}. \bibinfo{year}{2019}, p.
  \bibinfo{pages}{6411--6420}.
%Type = Inproceedings
\bibitem[{Meng et~al.(2019)Meng, Gao, Lai and Manocha}]{vvnet}
\bibinfo{author}{Meng\xfnm[ HY]}, \bibinfo{author}{Gao\xfnm[ L]},
  \bibinfo{author}{Lai\xfnm[ YK]}, \bibinfo{author}{Manocha\xfnm[ D]}.
\newblock \bibinfo{title}{Vv-net: Voxel vae net with group convolutions for
  point cloud segmentation}.
\newblock In: \bibinfo{booktitle}{Proceedings of the IEEE International
  Conference on Computer Vision}. \bibinfo{year}{2019}, p.
  \bibinfo{pages}{8500--8508}.
%Type = Article
\bibitem[{Hermosilla et~al.(2018)Hermosilla, Ritschel, V{\'a}zquez, Vinacua and
  Ropinski}]{montecarlo}
\bibinfo{author}{Hermosilla\xfnm[ P]}, \bibinfo{author}{Ritschel\xfnm[ T]},
  \bibinfo{author}{V{\'a}zquez\xfnm[ PP]}, \bibinfo{author}{Vinacua\xfnm[
  {\`A}]}, \bibinfo{author}{Ropinski\xfnm[ T]}.
\newblock \bibinfo{title}{Monte carlo convolution for learning on non-uniformly
  sampled point clouds}.
\newblock \bibinfo{journal}{ACM Transactions on Graphics (TOG)}
  \bibinfo{year}{2018};\bibinfo{volume}{37}(\bibinfo{number}{6}):\bibinfo{pages}{1--12}.
%Type = Inproceedings
\bibitem[{Goodfellow et~al.(2014)Goodfellow, Pouget-Abadie, Mirza, Xu,
  Warde-Farley, Ozair et~al.}]{goodfellow2014generative}
\bibinfo{author}{Goodfellow\xfnm[ I]}, \bibinfo{author}{Pouget-Abadie\xfnm[
  J]}, \bibinfo{author}{Mirza\xfnm[ M]}, \bibinfo{author}{Xu\xfnm[ B]},
  \bibinfo{author}{Warde-Farley\xfnm[ D]}, \bibinfo{author}{Ozair\xfnm[ S]},
  et~al.
\newblock \bibinfo{title}{Generative adversarial nets}.
\newblock In: \bibinfo{booktitle}{Advances in neural information processing
  systems (NIPS)}. \bibinfo{year}{2014}, p. \bibinfo{pages}{2672--2680}.
%Type = Inproceedings
\bibitem[{Arjovsky et~al.(2017)Arjovsky, Chintala and
  Bottou}]{arjovsky2017wasserstein}
\bibinfo{author}{Arjovsky\xfnm[ M]}, \bibinfo{author}{Chintala\xfnm[ S]},
  \bibinfo{author}{Bottou\xfnm[ L]}.
\newblock \bibinfo{title}{Wasserstein generative adversarial networks}.
\newblock In: \bibinfo{booktitle}{Proceedings of the 34th International
  Conference on Machine Learning-Volume 70}. \bibinfo{year}{2017}, p.
  \bibinfo{pages}{214--223}.
%Type = Inproceedings
\bibitem[{Kingma and Welling(2014)}]{kingma2013auto}
\bibinfo{author}{Kingma\xfnm[ D]}, \bibinfo{author}{Welling\xfnm[ M]}.
\newblock \bibinfo{title}{Auto-encoding variational bayes}.
\newblock In: \bibinfo{booktitle}{International Conference on Learning
  Representations (ICLR)}. \bibinfo{year}{2014},.
%Type = Inproceedings
\bibitem[{Achlioptas et~al.(2018)Achlioptas, Diamanti, Mitliagkas and
  Guibas}]{achlioptas2017latent_pc}
\bibinfo{author}{Achlioptas\xfnm[ P]}, \bibinfo{author}{Diamanti\xfnm[ O]},
  \bibinfo{author}{Mitliagkas\xfnm[ I]}, \bibinfo{author}{Guibas\xfnm[ L]}.
\newblock \bibinfo{title}{Learning representations and generative models for 3d
  point clouds}.
\newblock In: \bibinfo{booktitle}{International Conference on Learning
  Representations (ICLR)}. \bibinfo{year}{2018},.
%Type = Inproceedings
\bibitem[{Yang et~al.(2019)Yang, Huang, Hao, Liu, Belongie and
  Hariharan}]{pointflow}
\bibinfo{author}{Yang\xfnm[ G]}, \bibinfo{author}{Huang\xfnm[ X]},
  \bibinfo{author}{Hao\xfnm[ Z]}, \bibinfo{author}{Liu\xfnm[ MY]},
  \bibinfo{author}{Belongie\xfnm[ S]}, \bibinfo{author}{Hariharan\xfnm[ B]}.
\newblock \bibinfo{title}{Pointflow: 3d point cloud generation with continuous
  normalizing flows}.
\newblock In: \bibinfo{booktitle}{Proceedings of the IEEE International
  Conference on Computer Vision}. \bibinfo{year}{2019}, p.
  \bibinfo{pages}{4541--4550}.
%Type = Inproceedings
\bibitem[{Dubrovina et~al.(2019)Dubrovina, Xia, Achlioptas, Shalah, Groscot and
  Guibas}]{dubrovina2019composite}
\bibinfo{author}{Dubrovina\xfnm[ A]}, \bibinfo{author}{Xia\xfnm[ F]},
  \bibinfo{author}{Achlioptas\xfnm[ P]}, \bibinfo{author}{Shalah\xfnm[ M]},
  \bibinfo{author}{Groscot\xfnm[ R]}, \bibinfo{author}{Guibas\xfnm[ LJ]}.
\newblock \bibinfo{title}{Composite shape modeling via latent space
  factorization}.
\newblock In: \bibinfo{booktitle}{Proceedings of the IEEE International
  Conference on Computer Vision}. \bibinfo{year}{2019}, p.
  \bibinfo{pages}{8140--8149}.
%Type = Inproceedings
\bibitem[{Schor et~al.(2019)Schor, Katzir, Zhang and
  Cohen-Or}]{Schor_2019_ICCV}
\bibinfo{author}{Schor\xfnm[ N]}, \bibinfo{author}{Katzir\xfnm[ O]},
  \bibinfo{author}{Zhang\xfnm[ H]}, \bibinfo{author}{Cohen-Or\xfnm[ D]}.
\newblock \bibinfo{title}{Componet: Learning to generate the unseen by part
  synthesis and composition}.
\newblock In: \bibinfo{booktitle}{The IEEE International Conference on Computer
  Vision (ICCV)}. \bibinfo{year}{2019},.
%Type = Article
\bibitem[{Li et~al.(2019)Li, Niu and Xu}]{li2019learning}
\bibinfo{author}{Li\xfnm[ J]}, \bibinfo{author}{Niu\xfnm[ C]},
  \bibinfo{author}{Xu\xfnm[ K]}.
\newblock \bibinfo{title}{Learning part generation and assembly for
  structure-aware shape synthesis}.
\newblock \bibinfo{journal}{arXiv preprint arXiv:190606693}
  \bibinfo{year}{2019};.
%Type = Article
\bibitem[{Wang et~al.(2018)Wang, Schor, Hu, Huang, Cohen-Or and Huang}]{G2L18}
\bibinfo{author}{Wang\xfnm[ H]}, \bibinfo{author}{Schor\xfnm[ N]},
  \bibinfo{author}{Hu\xfnm[ R]}, \bibinfo{author}{Huang\xfnm[ H]},
  \bibinfo{author}{Cohen-Or\xfnm[ D]}, \bibinfo{author}{Huang\xfnm[ H]}.
\newblock \bibinfo{title}{Global-to-local generative model for 3d shapes}.
\newblock \bibinfo{journal}{ACM Transactions on Graphics (Proc SIGGRAPH ASIA)}
  \bibinfo{year}{2018};\bibinfo{volume}{37}(\bibinfo{number}{6}):\bibinfo{pages}{214:1—214:10}.
%Type = Inproceedings
\bibitem[{Shu et~al.(2019)Shu, Park and Kwon}]{treegan}
\bibinfo{author}{Shu\xfnm[ DW]}, \bibinfo{author}{Park\xfnm[ SW]},
  \bibinfo{author}{Kwon\xfnm[ J]}.
\newblock \bibinfo{title}{3d point cloud generative adversarial network based
  on tree structured graph convolutions}.
\newblock In: \bibinfo{booktitle}{Proceedings of the IEEE International
  Conference on Computer Vision}. \bibinfo{year}{2019}, p.
  \bibinfo{pages}{3859--3868}.
%Type = Article
\bibitem[{Mo et~al.(2019{\natexlab{a}})Mo, Guerrero, Yi, Su, Wonka, Mitra
  et~al.}]{Mo2019StructureNetHG}
\bibinfo{author}{Mo\xfnm[ K]}, \bibinfo{author}{Guerrero\xfnm[ P]},
  \bibinfo{author}{Yi\xfnm[ L]}, \bibinfo{author}{Su\xfnm[ H]},
  \bibinfo{author}{Wonka\xfnm[ P]}, \bibinfo{author}{Mitra\xfnm[ N]}, et~al.
\newblock \bibinfo{title}{Structurenet: Hierarchical graph networks for 3d
  shape generation}.
\newblock \bibinfo{journal}{ACM Trans Graph}
  \bibinfo{year}{2019}{\natexlab{a}};\bibinfo{volume}{38}:\bibinfo{pages}{242:1--242:19}.
%Type = Inproceedings
\bibitem[{Mo et~al.(2020)Mo, Guerrero, Yi, Su, Wonka, Mitra
  et~al.}]{structedit}
\bibinfo{author}{Mo\xfnm[ K]}, \bibinfo{author}{Guerrero\xfnm[ P]},
  \bibinfo{author}{Yi\xfnm[ L]}, \bibinfo{author}{Su\xfnm[ H]},
  \bibinfo{author}{Wonka\xfnm[ P]}, \bibinfo{author}{Mitra\xfnm[ NJ]}, et~al.
\newblock \bibinfo{title}{Structedit: Learning structural shape variations}.
\newblock In: \bibinfo{booktitle}{Proceedings of the IEEE/CVF Conference on
  Computer Vision and Pattern Recognition}. \bibinfo{year}{2020}, p.
  \bibinfo{pages}{8859--8868}.
%Type = Article
\bibitem[{Gao et~al.(2019)Gao, Yang, Wu, Yuan, Fu, Lai et~al.}]{sdmnet}
\bibinfo{author}{Gao\xfnm[ L]}, \bibinfo{author}{Yang\xfnm[ J]},
  \bibinfo{author}{Wu\xfnm[ T]}, \bibinfo{author}{Yuan\xfnm[ YJ]},
  \bibinfo{author}{Fu\xfnm[ H]}, \bibinfo{author}{Lai\xfnm[ YK]}, et~al.
\newblock \bibinfo{title}{Sdm-net: Deep generative network for structured
  deformable mesh}.
\newblock \bibinfo{journal}{ACM Transactions on Graphics (TOG)}
  \bibinfo{year}{2019};\bibinfo{volume}{38}(\bibinfo{number}{6}):\bibinfo{pages}{1--15}.
%Type = Article
\bibitem[{Yin et~al.(2020)Yin, Chen, Chaudhuri, Fisher, Kim and
  Zhang}]{coalesce}
\bibinfo{author}{Yin\xfnm[ K]}, \bibinfo{author}{Chen\xfnm[ Z]},
  \bibinfo{author}{Chaudhuri\xfnm[ S]}, \bibinfo{author}{Fisher\xfnm[ M]},
  \bibinfo{author}{Kim\xfnm[ V]}, \bibinfo{author}{Zhang\xfnm[ H]}.
\newblock \bibinfo{title}{Coalesce: Component assembly by learning to
  synthesize connections}.
\newblock \bibinfo{journal}{arXiv preprint arXiv:200801936}
  \bibinfo{year}{2020};.
%Type = Inproceedings
\bibitem[{Gulrajani et~al.(2017)Gulrajani, Ahmed, Arjovsky, Dumoulin and
  Courville}]{gulrajani2017improved}
\bibinfo{author}{Gulrajani\xfnm[ I]}, \bibinfo{author}{Ahmed\xfnm[ F]},
  \bibinfo{author}{Arjovsky\xfnm[ M]}, \bibinfo{author}{Dumoulin\xfnm[ V]},
  \bibinfo{author}{Courville\xfnm[ AC]}.
\newblock \bibinfo{title}{Improved training of wasserstein gans}.
\newblock In: \bibinfo{booktitle}{Advances in neural information processing
  systems}. \bibinfo{year}{2017}, p. \bibinfo{pages}{5767--5777}.
%Type = Article
\bibitem[{Higgins et~al.(2017)Higgins, Matthey, Pal, Burgess, Glorot, Botvinick
  et~al.}]{higgins2017beta}
\bibinfo{author}{Higgins\xfnm[ I]}, \bibinfo{author}{Matthey\xfnm[ L]},
  \bibinfo{author}{Pal\xfnm[ A]}, \bibinfo{author}{Burgess\xfnm[ C]},
  \bibinfo{author}{Glorot\xfnm[ X]}, \bibinfo{author}{Botvinick\xfnm[ M]},
  et~al.
\newblock \bibinfo{title}{beta-vae: Learning basic visual concepts with a
  constrained variational framework.}
\newblock \bibinfo{journal}{International Conference on Learning
  Representations (ICLR)}
  \bibinfo{year}{2017};\bibinfo{volume}{2}(\bibinfo{number}{5}):\bibinfo{pages}{6}.
%Type = Article
\bibitem[{Kullback and Leibler(1951)}]{kullback1951}
\bibinfo{author}{Kullback\xfnm[ S]}, \bibinfo{author}{Leibler\xfnm[ RA]}.
\newblock \bibinfo{title}{On information and sufficiency}.
\newblock \bibinfo{journal}{Ann Math Statist}
  \bibinfo{year}{1951};\bibinfo{volume}{22}(\bibinfo{number}{1}):\bibinfo{pages}{79--86}.
\newblock \DOIprefix\doi{10.1214/aoms/1177729694}.
%Type = Article
\bibitem[{Yi et~al.(2016)Yi, Kim, Ceylan, Shen, Yan, Su et~al.}]{Yi16}
\bibinfo{author}{Yi\xfnm[ L]}, \bibinfo{author}{Kim\xfnm[ VG]},
  \bibinfo{author}{Ceylan\xfnm[ D]}, \bibinfo{author}{Shen\xfnm[ IC]},
  \bibinfo{author}{Yan\xfnm[ M]}, \bibinfo{author}{Su\xfnm[ H]}, et~al.
\newblock \bibinfo{title}{A scalable active framework for region annotation in
  3d shape collections}.
\newblock \bibinfo{journal}{SIGGRAPH Asia} \bibinfo{year}{2016};.
%Type = Techreport
\bibitem[{Chang et~al.(2015)Chang, Funkhouser, Guibas, Hanrahan, Huang, Li
  et~al.}]{shapenet2015}
\bibinfo{author}{Chang\xfnm[ AX]}, \bibinfo{author}{Funkhouser\xfnm[ T]},
  \bibinfo{author}{Guibas\xfnm[ L]}, \bibinfo{author}{Hanrahan\xfnm[ P]},
  \bibinfo{author}{Huang\xfnm[ Q]}, \bibinfo{author}{Li\xfnm[ Z]}, et~al.
\newblock \bibinfo{title}{{ShapeNet: An Information-Rich 3D Model Repository}}.
\newblock \bibinfo{type}{Tech. Rep.} \bibinfo{number}{arXiv:1512.03012
  [cs.GR]}; Stanford University --- Princeton University --- Toyota
  Technological Institute at Chicago; \bibinfo{year}{2015}.
%Type = Article
\bibitem[{Ravi et~al.(2020)Ravi, Reizenstein, Novotny, Gordon, Lo, Johnson
  et~al.}]{ravi2020pytorch3d}
\bibinfo{author}{Ravi\xfnm[ N]}, \bibinfo{author}{Reizenstein\xfnm[ J]},
  \bibinfo{author}{Novotny\xfnm[ D]}, \bibinfo{author}{Gordon\xfnm[ T]},
  \bibinfo{author}{Lo\xfnm[ WY]}, \bibinfo{author}{Johnson\xfnm[ J]}, et~al.
\newblock \bibinfo{title}{Accelerating 3d deep learning with pytorch3d}.
\newblock \bibinfo{journal}{arXiv:200708501} \bibinfo{year}{2020};.
%Type = Inproceedings
\bibitem[{{Fan} et~al.(2017){Fan}, {Su} and {Guibas}}]{fan2017point}
\bibinfo{author}{{Fan}\xfnm[ H]}, \bibinfo{author}{{Su}\xfnm[ H]},
  \bibinfo{author}{{Guibas}\xfnm[ L]}.
\newblock \bibinfo{title}{A point set generation network for 3d object
  reconstruction from a single image}.
\newblock In: \bibinfo{booktitle}{2017 IEEE Conference on Computer Vision and
  Pattern Recognition (CVPR)}. \bibinfo{year}{2017}, p.
  \bibinfo{pages}{2463--2471}.
\newblock \DOIprefix\doi{10.1109/CVPR.2017.264}.
%Type = Article
\bibitem[{Rubner et~al.(2000)Rubner, Tomasi and Guibas}]{rubner2000earth}
\bibinfo{author}{Rubner\xfnm[ Y]}, \bibinfo{author}{Tomasi\xfnm[ C]},
  \bibinfo{author}{Guibas\xfnm[ LJ]}.
\newblock \bibinfo{title}{The earth mover's distance as a metric for image
  retrieval}.
\newblock \bibinfo{journal}{International Journal of Computer Vision}
  \bibinfo{year}{2000};\bibinfo{volume}{40}(\bibinfo{number}{2}):\bibinfo{pages}{99--121}.
%Type = Inproceedings
\bibitem[{Kingma and Ba(2015)}]{kingma2014adam}
\bibinfo{author}{Kingma\xfnm[ DP]}, \bibinfo{author}{Ba\xfnm[ J]}.
\newblock \bibinfo{title}{Adam: {A} method for stochastic optimization}.
\newblock In: \bibinfo{booktitle}{3rd International Conference on Learning
  Representations, {ICLR} 2015, San Diego, CA, USA, May 7-9, 2015, Conference
  Track Proceedings}. \bibinfo{year}{2015},.
%Type = Inproceedings
\bibitem[{Wu et~al.(2020)Wu, Chen, Zhuang and Chen}]{wu_2020_ECCV}
\bibinfo{author}{Wu\xfnm[ R]}, \bibinfo{author}{Chen\xfnm[ X]},
  \bibinfo{author}{Zhuang\xfnm[ Y]}, \bibinfo{author}{Chen\xfnm[ B]}.
\newblock \bibinfo{title}{Multimodal shape completion via conditional
  generative adversarial networks}.
\newblock In: \bibinfo{booktitle}{The European Conference on Computer Vision
  (ECCV)}. \bibinfo{year}{2020},.
%Type = Inproceedings
\bibitem[{Mo et~al.(2019{\natexlab{b}})Mo, Zhu, Chang, Yi, Tripathi, Guibas
  et~al.}]{partnet}
\bibinfo{author}{Mo\xfnm[ K]}, \bibinfo{author}{Zhu\xfnm[ S]},
  \bibinfo{author}{Chang\xfnm[ AX]}, \bibinfo{author}{Yi\xfnm[ L]},
  \bibinfo{author}{Tripathi\xfnm[ S]}, \bibinfo{author}{Guibas\xfnm[ LJ]},
  et~al.
\newblock \bibinfo{title}{{PartNet}: A large-scale benchmark for fine-grained
  and hierarchical part-level {3D} object understanding}.
\newblock In: \bibinfo{booktitle}{The IEEE Conference on Computer Vision and
  Pattern Recognition (CVPR)}. \bibinfo{year}{2019}{\natexlab{b}},.

\end{thebibliography}
\newpage
\section*{Appendix}

\section{Ablation Study}
\label{sec:apablation}

\begin{table}
\caption{The effect of different variations of feature extractor and segmentation module based on reconstruction loss (Chamfer) and segmentation accuracy. }
\label{table:variations}
\begin{tabular}{l c c c c}
 & \multicolumn{2}{c}{Rec. loss ($\times10^{-4}$)} & \multicolumn{2}{c}{Seg. acc. \%} \\
 & Train & Test & Train & Test \\
\hline
Base Model & 3.61 & 5.93 & 96.23 & 93.51 \\
\hline
\multicolumn{5}{c}{Feature Extractor} \\
\hline
PointNet \cite{qi2016pointnet} & 3.92 & 6.06 & 96.35 & 93.84 \\
Mean pooling & 5.48 & 7.01 & 96.18 & 93.61 \\
\hline
\multicolumn{5}{c}{Segmentation module} \\
\hline
No module & 3.01 & 5.20 & - & - \\
Module failure & 3.11 & 5.95 & - & - \\
No global features & 4.24 & 6.04 & 87.47 & 86.95 \\

\hline
\end{tabular}
\end{table}

The proposed framework allows replacement of the Feature extractor and Segmentation modules. Table \ref{table:variations} summarizes the reconstruction and segmentation performance by (\textit{i}) substituting feature extraction with PointNet while keeping the other modules the same and changing its pooling layer with mean pooling; (\textit{ii}) experimenting on segmentation module by removing it, using a sub-optimal segmentation module and using a segmentation module omitting the global features. All variations are trained with the same parameters.

\begin{figure}
\centering
\rotatebox{90}{\hspace{0.2cm}Test set}
\includegraphics[width = 0.46\textwidth]{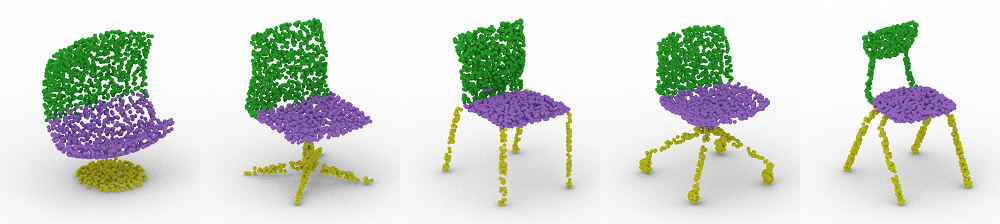}\\
\rotatebox{90}{\hspace{0.5cm}Max}
\includegraphics[width = 0.46\textwidth]{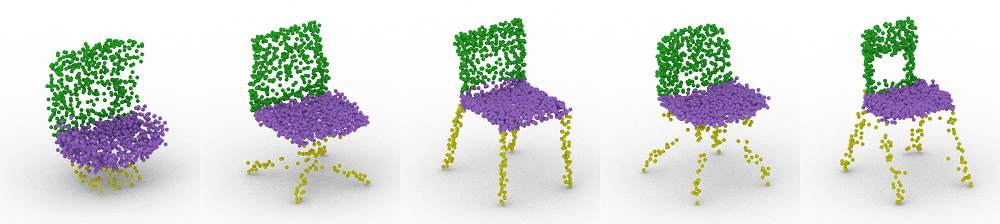}\\
\rotatebox{90}{\hspace{0.5cm}Mean}
\includegraphics[width = 0.46\textwidth]{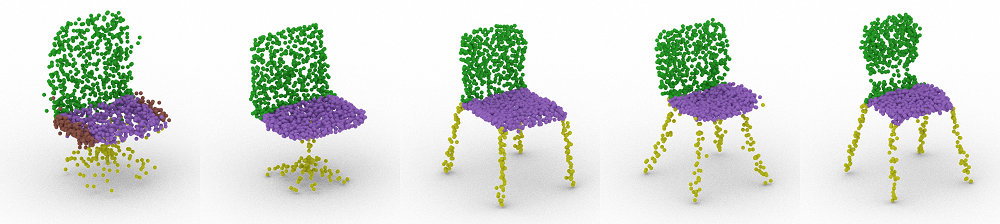}
\caption{Reconstruction comparison between models using max and mean pooling.}
\label{fig:apmeanpooling}
\end{figure}

Replacing the feature extraction with PointNet does not provide any benefits since the samples are already aligned and the system works with a single class. Since the input data has a single class and limited diversity, a 3-layer model is sufficient for extracting the necessary features and using 5-layers does not provide any advantage. Replacing the max-pooling with mean-pooling, which is also a symmetric operation, degrades the results. Mean-pooling extracts the average of features rather than selecting the most effective and critical features like max-pooling. The extracted features represent an average model with smooth edges and it causes poor reconstructions for complex and unusual models which can be seen in Fig. \ref{fig:apmeanpooling}.

\begin{figure*}
\centering
\includegraphics[width = 0.12\textwidth]{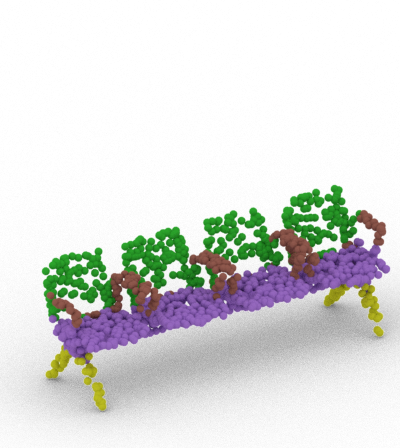}
\includegraphics[width = 0.12\textwidth]{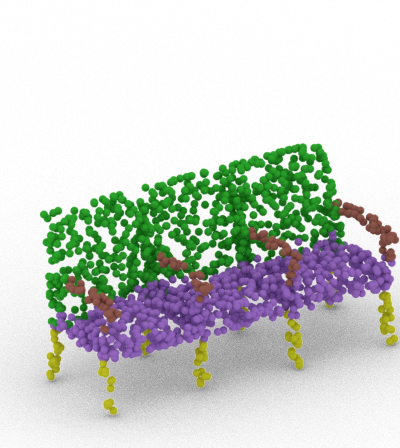}
\includegraphics[width = 0.12\textwidth]{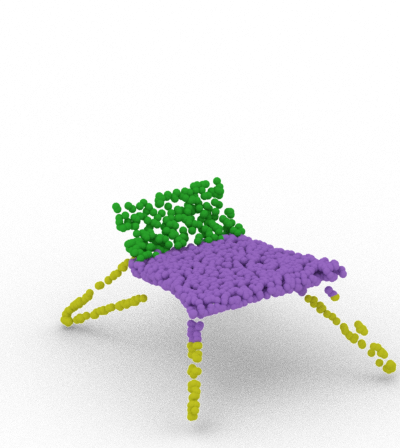}
\includegraphics[width = 0.12\textwidth]{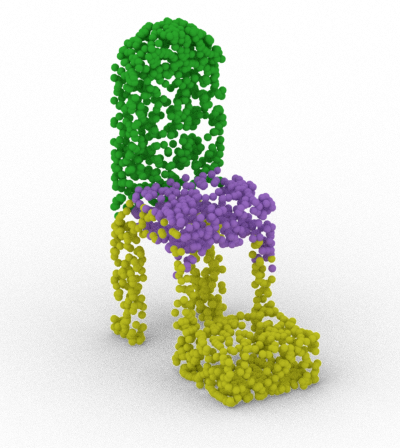}
\includegraphics[width = 0.12\textwidth]{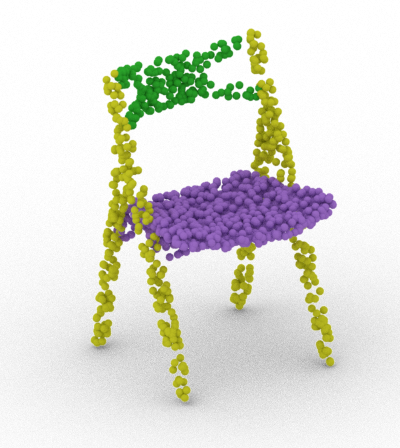}
\includegraphics[width = 0.12\textwidth]{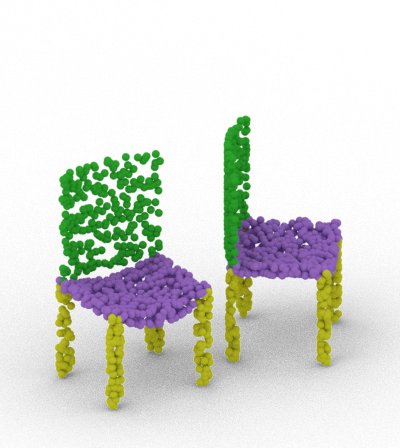}
\includegraphics[width = 0.12\textwidth]{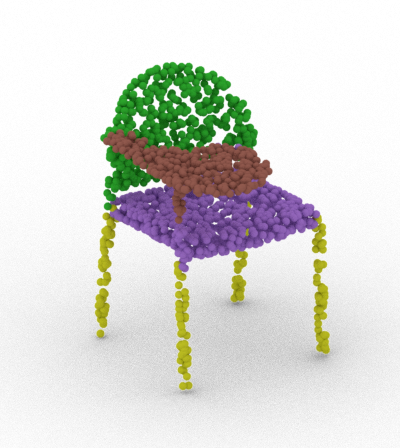}
\includegraphics[width = 0.12\textwidth]{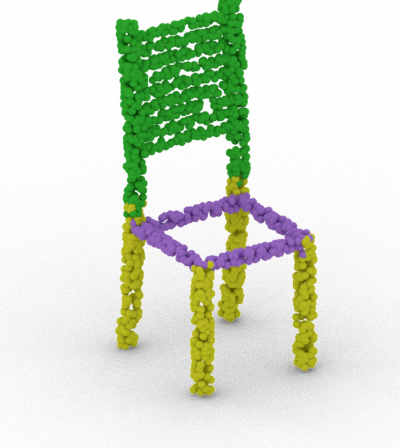}\\
\includegraphics[width = 0.12\textwidth]{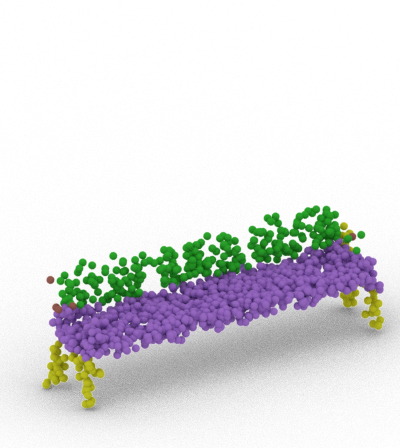}
\includegraphics[width = 0.12\textwidth]{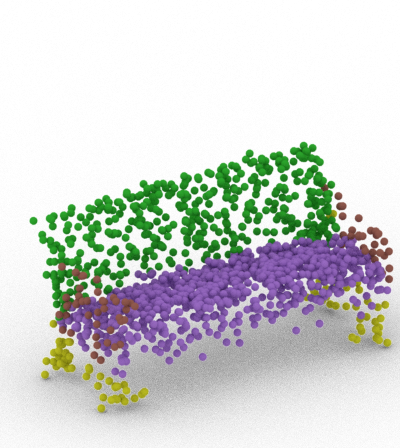}
\includegraphics[width = 0.12\textwidth]{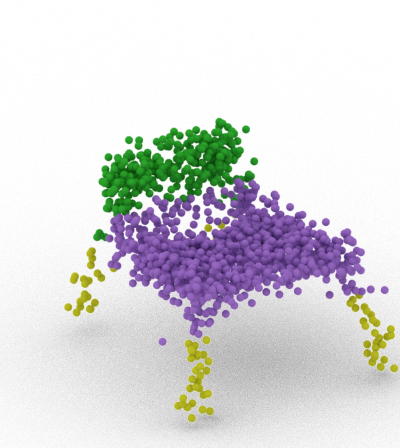}
\includegraphics[width = 0.12\textwidth]{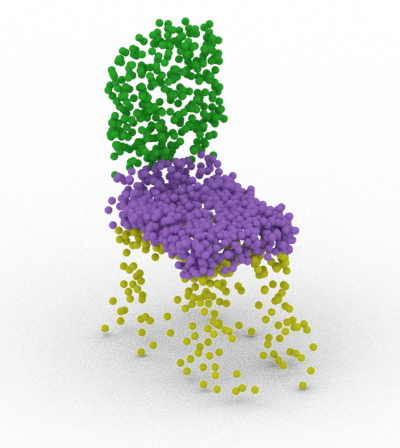}
\includegraphics[width = 0.12\textwidth]{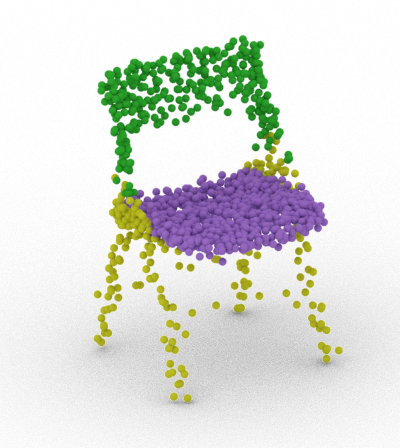}
\includegraphics[width = 0.12\textwidth]{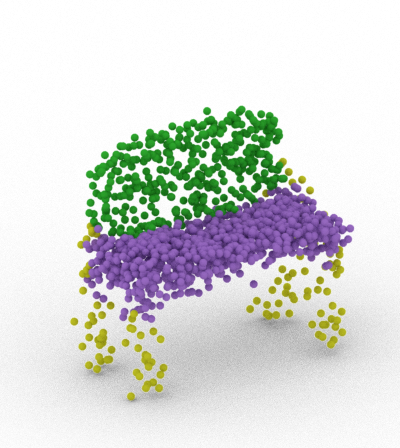}
\includegraphics[width = 0.12\textwidth]{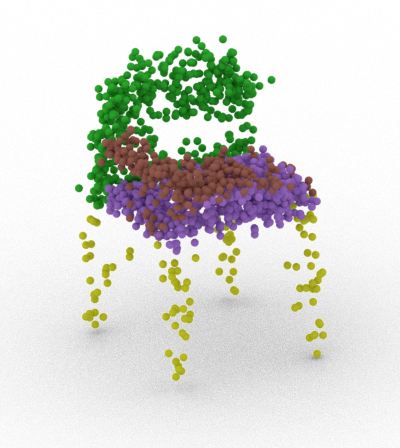}
\includegraphics[width = 0.12\textwidth]{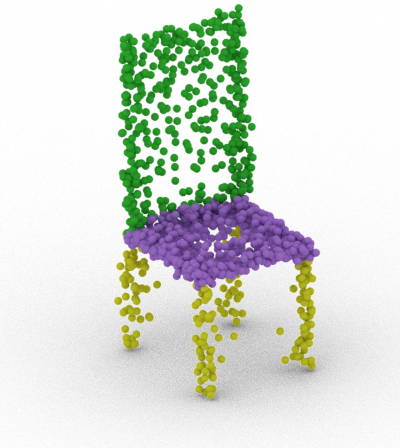}\\
 (a) \hspace{0.093\textwidth} (b) \hspace{0.093\textwidth} (c) \hspace{0.093\textwidth} (d) \hspace{0.093\textwidth} (e) \hspace{0.093\textwidth} (f) \hspace{0.093\textwidth} (g) \hspace{0.093\textwidth} (h)
\caption{The reconstruction results of failure cases. The failures are mostly the result of outliers such as unusual and asymmetric cases.}
\label{fig:apfailrec}
\end{figure*}

To observe the effect of the segmentation module, we trained the system without it and fed the ground truth part labels. Since the part labels are not predictions but ground truths, the reconstruction performance was better as expected. On the other hand, elimination of segmentation module results in an undesirable effect of eliminating the ability of the system to work with unannotated raw point clouds. We also deliberately hindered the training of segmentation module and randomly initialized the module to simulate segmentation failures where segmentation results are random. Interestingly, it is quantitatively better than the base model because now each point is randomly assigned to different parts, thus each part is simply a downsampled version of the global model. All part features are equal to global feature so the system captures the global features better. However, in this case, the system does not have any part-based abilities anymore and not fit for purpose since all parts are equivalent to global model. Lastly, the segmentation module in the base system is trained with only point features (without concatenating global features). Lower segmentation performance highlights the importance of global features -alongside point features- in the segmentation performance. 

\section{Failure Cases}
\label{sec:apfailures}

The samples with high reconstruction losses are visualized to analyze failure cases in Fig. \ref{fig:apfailrec} where the test samples are shown at the top row and their respective reconstructions at the bottom row. The unusual object samples in Fig. \ref{fig:apfailrec} (a)-(d) are outliers and their respective reconstructions are noisy. Chairs in Fig. \ref{fig:apfailrec} (a) and (b) have arms in the middle, this is not a common occurrence in the training set and these arms could not be represented. Unusual leg shapes of chairs in Fig. \ref{fig:apfailrec} (c) and (d) can not be reconstructed well, resulting in high reconstruction loss. Reconstructed part labels are different for Fig. \ref{fig:apfailrec} (e) due to segmentation error. However, the reconstructed shape is still acceptable because leg and back parts are ambiguously defined. Chairs in Fig. \ref{fig:apfailrec} (f) and (g) have highly asymmetric shapes. Asymmetric shapes comprise less than 3\% of the whole dataset, so the system can not adequately learn to represent them. This can be prevented by augmenting the dataset with further asymmetric samples. However, to generate novel asymmetric structures, more explicit constraints must be defined. Fig. \ref{fig:apfailrec} (h) has an unusual hole on the seat part, again not present in the training set. All of the reconstruction errors are because of the lack of representative samples in the training set and can be prevented by extending the dataset with more diverse samples.

\begin{figure}
\centering
\includegraphics[width = 0.1\textwidth]{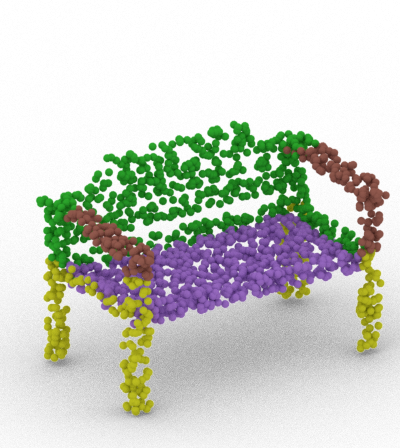} \hfill
\includegraphics[width = 0.1\textwidth]{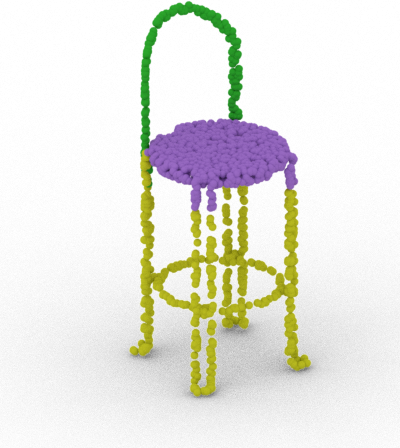}\\
\includegraphics[width = 0.48\textwidth]{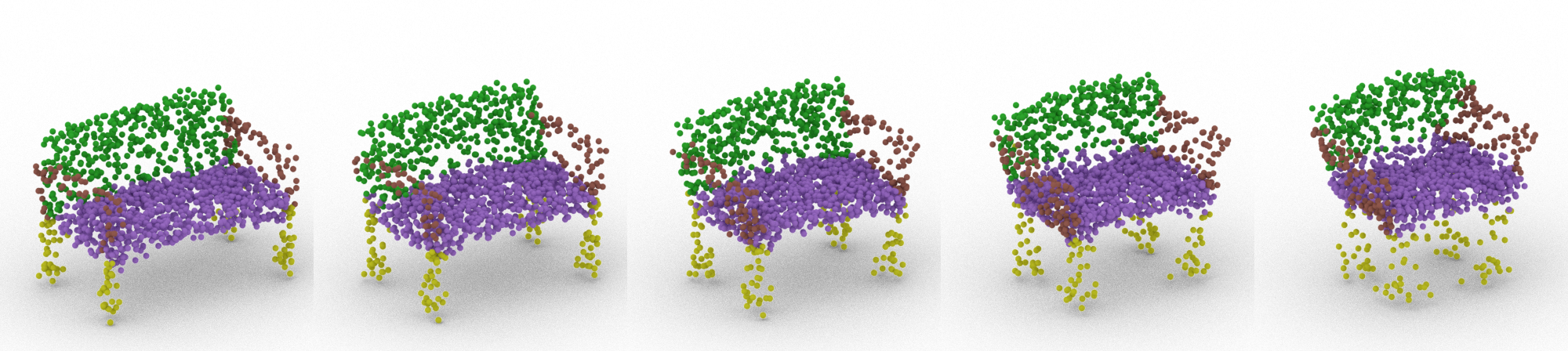}
\caption{A challenging part (leg) interpolation between two distant shapes.}
\label{fig:apfailinterp}
\end{figure}

Interpolation between distant shapes such as the ones in Fig. \ref{fig:apfailinterp} may not be successful. On the other hand, it can be argued that, this operation is hardly plausible for humans as well. The target leg on the right is not easy to seamlessly merge into the original global shape on the left and the model does its best to modify the source shape and leg to generate a semantically acceptable global shape. This supports our claim that the system makes semantic modifications. However, it cannot be loyal to the original shapes for this case as this would interfere with generating a semantically acceptable global shape.

\section{Further Experiments}

\subsection{Supplementary Comparisons}
\label{sec:apcomp}
 StructureNet \cite{Mo2019StructureNetHG} is designed to work on a fine-grained dataset hierarchically labeled with child parts such as PartNet \cite{partnet}. This dataset structure is fundamentally different to the one we used in this work. So, in order to facilitate comparisons, we have also trained our model with the PartNet where each part and child parts have 1000 points. We have grouped all the child parts into the same semantic definitions as we used such as seat, back, leg and arm. Both models have been trained and evaluated using CD. The results are reported for chair class, which has 4871 samples divided with 7:1:2 ratio for training, validation and test respectively, with 2048 points per sample. 

\begin{figure}
\centering
\rotatebox{90}{\hspace{0.2cm}Test set}
\includegraphics[width = 0.46\textwidth]{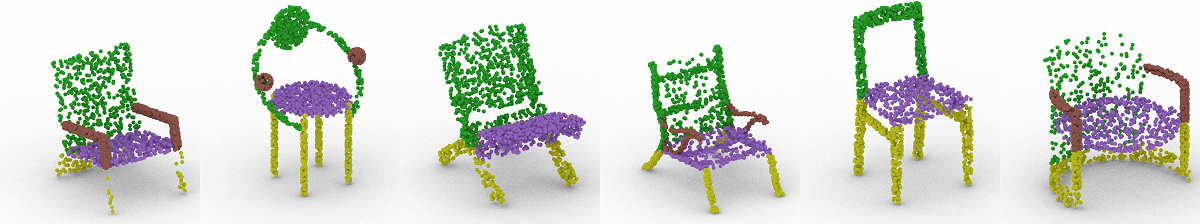}\\
\rotatebox{90}{StructureNet}
\includegraphics[width = 0.46\textwidth]{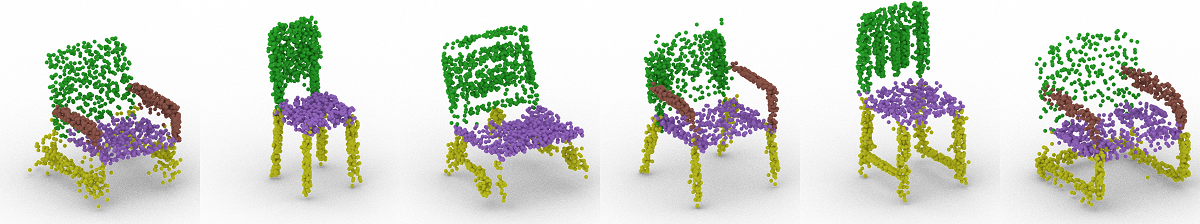}\\
\rotatebox{90}{\hspace{0.4cm}Ours}
\includegraphics[width = 0.46\textwidth]{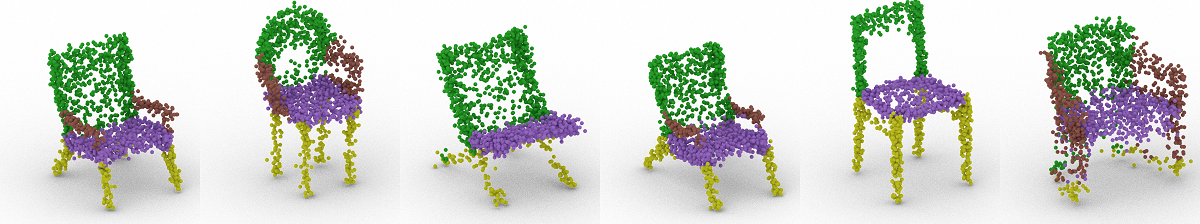}
\caption{Reconstruction results of challenging cases for StructureNet\cite{Mo2019StructureNetHG} and our model.}
\label{fig:apstnetrecon}
\end{figure}

The average reconstruction error (Chamfer$\times10^{-4}$) for the global shapes is calculated as $30.18$ for StructureNet and $12.11$ for our model. The reconstruction results are similar for common cases but the results for challenging cases can be seen in Fig. \ref{fig:apstnetrecon}. Our results become noisy but represent the global shape better. The noise in StructureNet appears as structural inaccuracies since it makes structural encoding-decoding. It is also reported that noise in StructureNet may result in missing parts, duplicate parts, detached parts \cite{Mo2019StructureNetHG}. Considering both quantitative and qualitative comparison, the proposed model performs better at global shape reconstruction. StructureNet generates novel structures and parts using VAE. We compared the new sample generation capabilities of both models with the evaluation metrics we used. The results provided in Table \ref{table:stnetgens} show that the proposed model has better MMD, Coverage and JSD scores.

\begin{table}
\small
\begin{center}
\caption{Comparison with StructureNet on the PartNet \cite{partnet}}
\label{table:stnetgens}
\begin{tabular}{l c c c}
Model & MMD &\% Cov & JSD \\
\hline
VAE & \textbf{17.27} & \textbf{67.96} & 20.61 \\
GAN & 30.74 & 24.21 & 19.71 \\ 
WGAN & 21.93 & 62.50 & \textbf{10.34} \\
StructureNet \cite{Mo2019StructureNetHG} & 27.14 & 39.06 & 17.98 \\
\hline
\end{tabular}
\end{center}
\end{table}

\begin{figure}
\rotatebox{90}{StructureNet}
\includegraphics[width=0.48\textwidth]{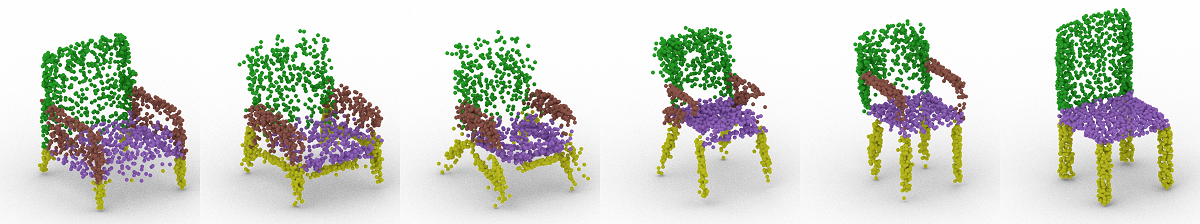}\\
\bigskip
\rotatebox{90}{\hspace{0.4cm} Ours}
\includegraphics[width=0.48\textwidth]{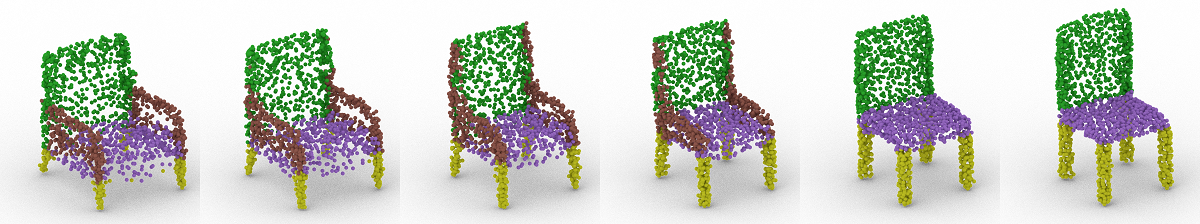}\\
\rotatebox{90}{CompoNet}
\includegraphics[width=0.48\textwidth]{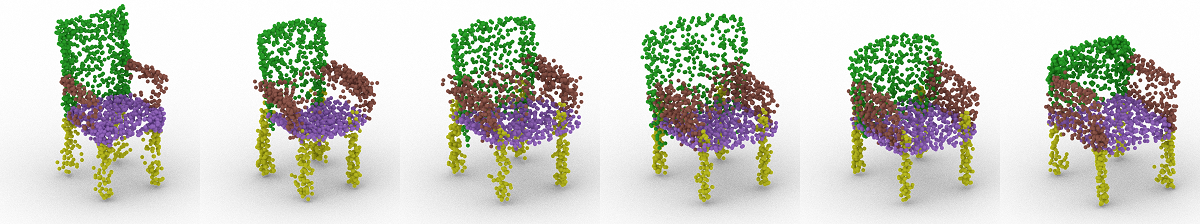}\\
\rotatebox{90}{\hspace{0.4cm} Ours}
\includegraphics[width=0.48\textwidth]{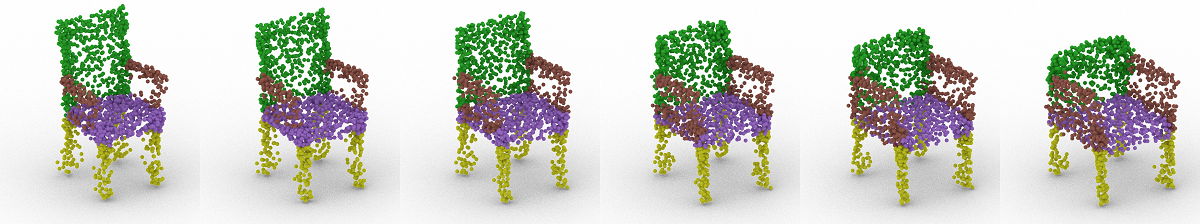}
\caption{Interpolation comparison with StructureNet \cite{Mo2019StructureNetHG} and CompoNet \cite{Schor_2019_ICCV}  between the same two shapes (leftmost and rightmost).}
\label{fig:apinterpcomp}
\end{figure}

Visual interpolation results for different methods are provided in Fig. \ref{fig:apinterpcomp}. As StructureNet performs structure interpolation by its nature, it causes sharp structural changes. CompoNet performs per-part interpolation, however it suffers from part assembly problems in some steps. In both cases, the proposed model performs a smooth global shape interpolation, generating plausible global shapes during the transition steps.

\subsection{TMD scores for other classes}
\label{sec:aptmd}
TMD results for \textit{table} and \textit{plane} classes can be found in Tables \ref{table:tmd_table} and \ref{table:tmd_plane} respectively. \textit{Table} class has higher TMD scores relative to \textit{chair} class since it has only 2 parts; top and leg. A missing part means half the parts of the model are missing and the completion causes higher diversity. \textit{Plane} class has lower TMD scores, implying less diversity. This is expected since the plane models are smaller, more dense, less diverse and they occupy a smaller space.

\begin{table}
\begin{center}
\caption{Total Mutual Difference (TMD$\times10^{-2}$) scores for \textit{table} class.}
\label{table:tmd_table}
\begin{tabular}{l c c}
\hline
 & \multicolumn{2}{c}{\# of changing parts} \\ 
Model & 1 & 2\\
\hline
Exchange & 5.27 & 9.89\\ 
VAE & 3.31 & 6.02\\
l-GAN & 3.27 & 4.64\\ 
l-WGAN & 3.57 & 6.95\\ 
\hline
\end{tabular}
\end{center}
\end{table}

\begin{table}[h]
\begin{center}
\caption{Total Mutual Difference (TMD$\times10^{-2}$) scores for \textit{plane} class.}
\label{table:tmd_plane}
\begin{tabular}{l c c c c}
\hline
 & \multicolumn{4}{c}{\# of changing parts} \\ 
Model & 1 & 2 & 3 & 4\\
\hline
Exchange & 0.23 & 1.00 & 1.11 & 1.15\\ 
VAE & 0.21 & 0.64 & 0.69 & 0.73\\
l-GAN & 0.13 & 0.28 & 0.33 & 0.36\\ 
l-WGAN & 0.21 & 0.69 & 0.77 & 0.80\\ 
\hline
\end{tabular}
\end{center}
\end{table}

\subsection{Visualizations of other classes}
\label{sec:apvis}
\begin{figure*}
\centering
\captionsetup{justification=centering}
\rotatebox{90}{\hspace{0cm} Global}
\includegraphics[width = 0.9\textwidth, trim=0 0 0 2cm, clip]{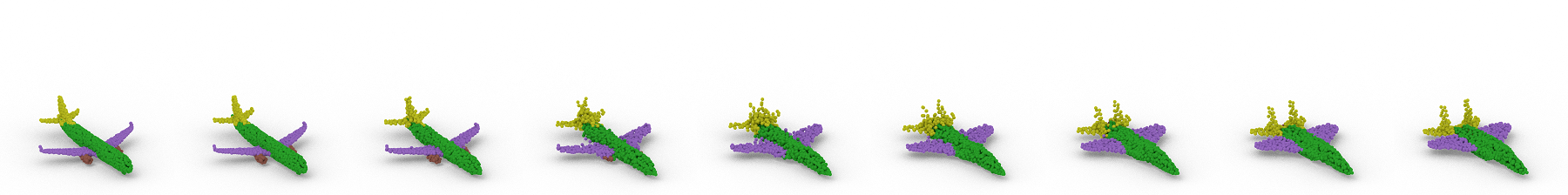}\\
\rotatebox{90}{\hspace{0cm} Body}
\includegraphics[width = 0.9\textwidth, trim=0 0 0 2cm, clip]{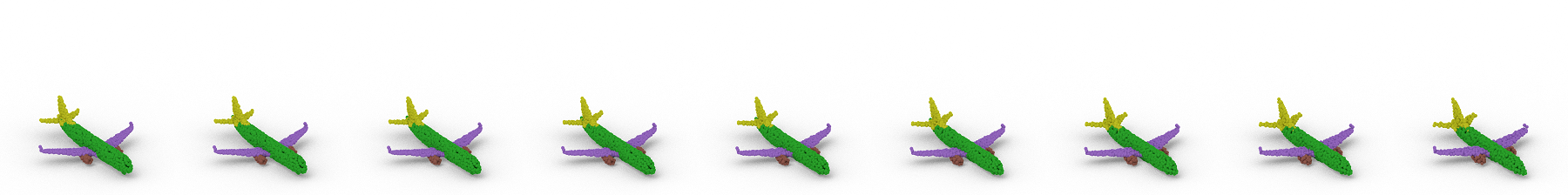}\\
\rotatebox{90}{\hspace{0cm} Wings}
\includegraphics[width = 0.9\textwidth, trim=0 0 0 2cm, clip]{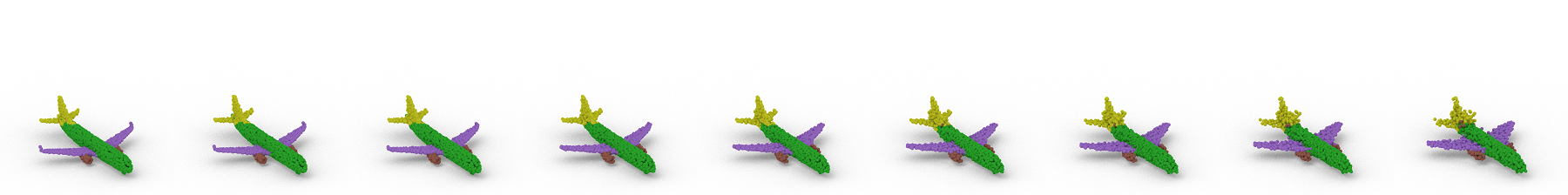}\\
\rotatebox{90}{\hspace{0.1cm} Tail}
\includegraphics[width = 0.9\textwidth, trim=0 0 0 2cm, clip]{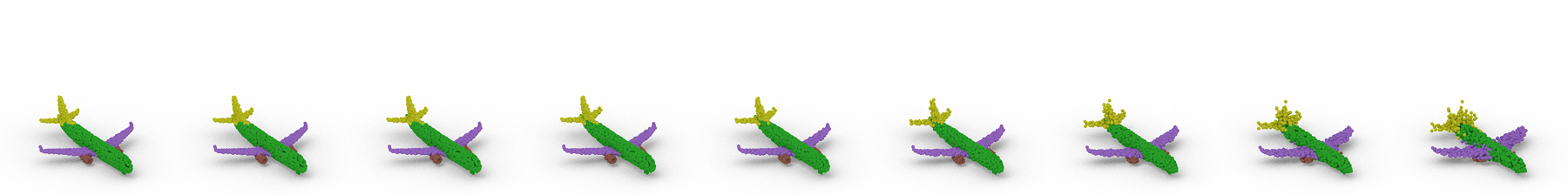}\\
\rotatebox{90}{\hspace{0cm} Engines}
\includegraphics[width = 0.9\textwidth, trim=0 0 0 2cm, clip]{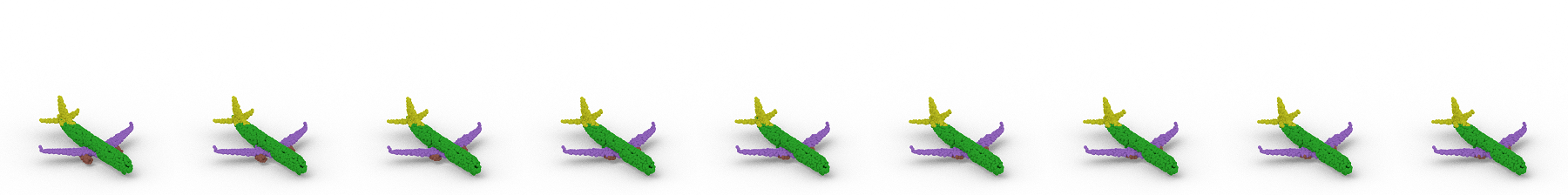}\\
\vspace{0.5cm}
\rotatebox{90}{\hspace{0.1cm} Global}
\includegraphics[width = 0.9\textwidth]{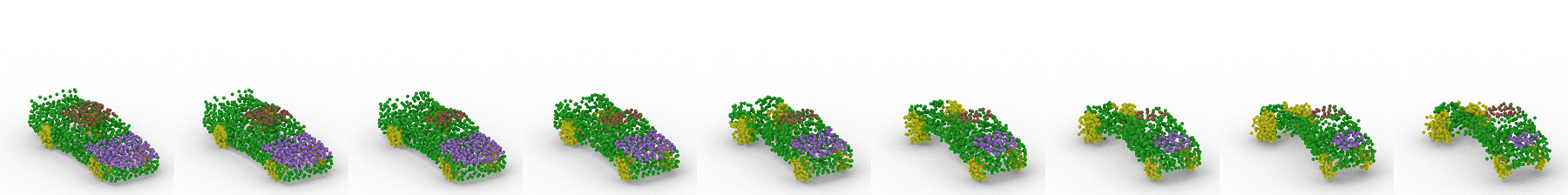}\\
\rotatebox{90}{\hspace{0.1cm} Hood}
\includegraphics[width = 0.9\textwidth]{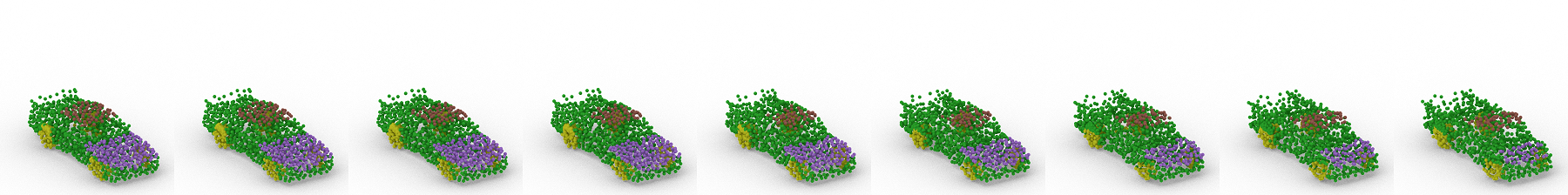}\\
\rotatebox{90}{\hspace{0.1cm} Roof}
\includegraphics[width = 0.9\textwidth]{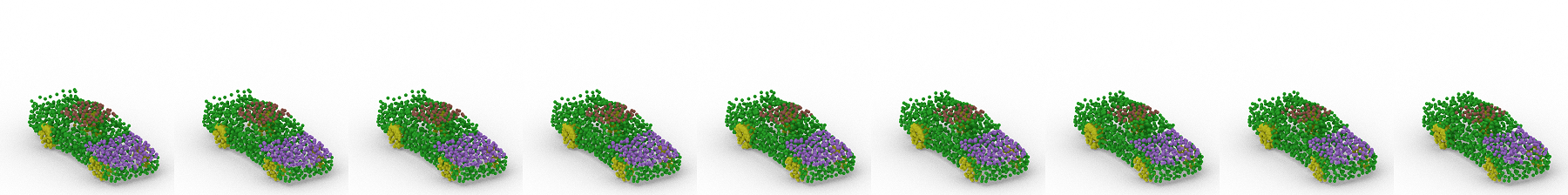}\\
\rotatebox{90}{\hspace{0.1cm} Wheels}
\includegraphics[width = 0.9\textwidth]{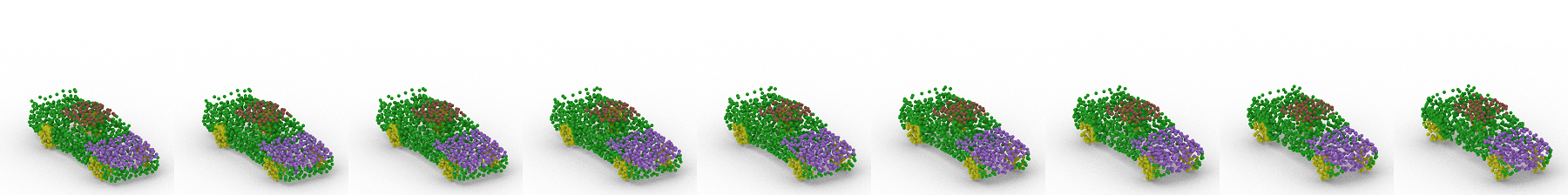}\\
\rotatebox{90}{\hspace{0.1cm} Body}
\includegraphics[width = 0.9\textwidth]{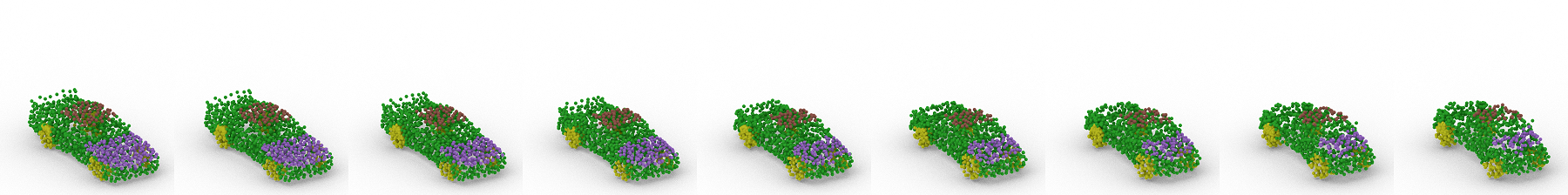}\\
\vspace{0.5cm}
\rotatebox{90}{\hspace{0.1cm} Global}
\includegraphics[width = 0.9\textwidth]{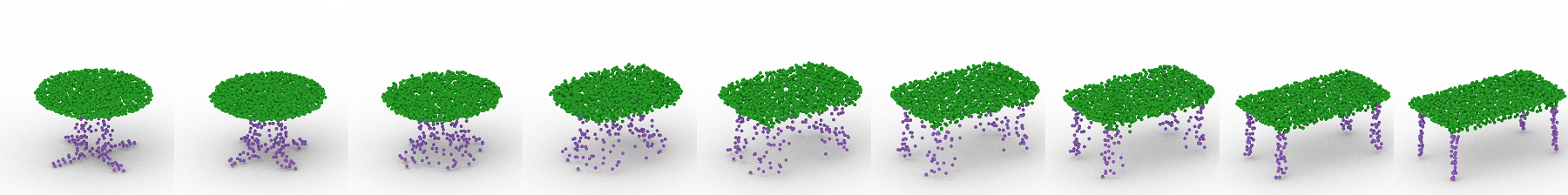}\\
\rotatebox{90}{\hspace{0.1cm} Top}
\includegraphics[width = 0.9\textwidth]{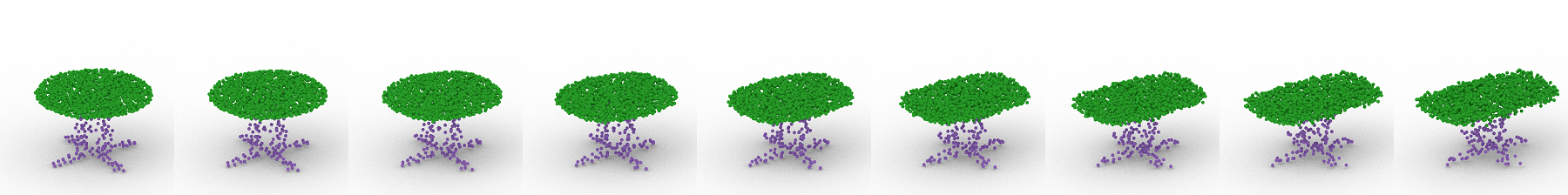}\\
\rotatebox{90}{\hspace{0.1cm} Foot}
\includegraphics[width = 0.9\textwidth]{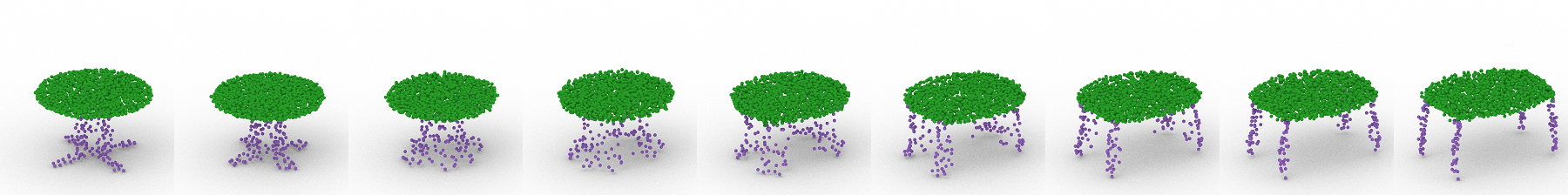}\\
\caption{Part interpolation results for plane, car and table classes.}
\label{fig:apinterp}
\end{figure*}

\begin{figure*}
\centering
\rotatebox{90}{\hspace{0cm} VAE}
\includegraphics[width = 0.95\textwidth, trim=0 0 0 1cm, clip]{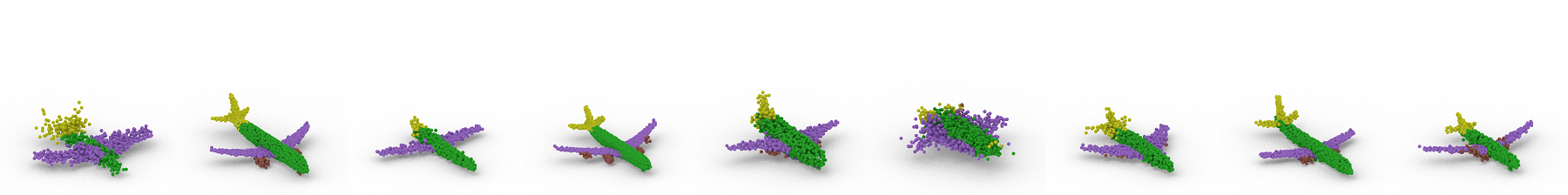}\\
\rotatebox{90}{\hspace{0cm} GAN}
\includegraphics[width = 0.95\textwidth, trim=0 0 0 1cm, clip]{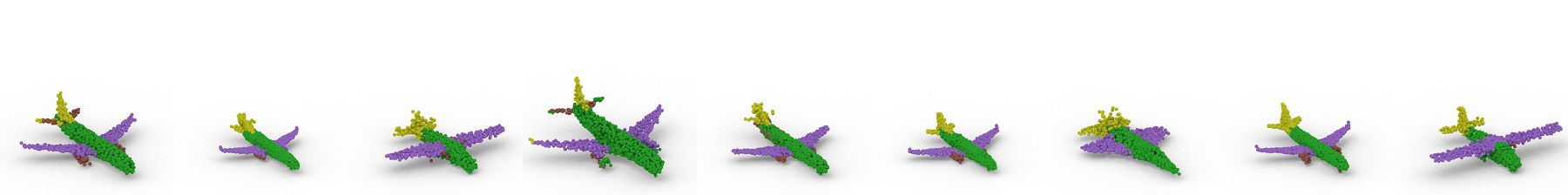}\\
\rotatebox{90}{\hspace{0cm} WGAN}
\includegraphics[width = 0.95\textwidth, trim=0 0 0 1cm, clip]{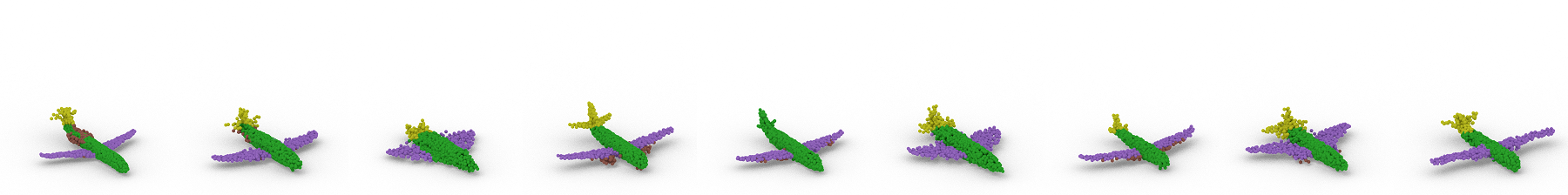}\\
\vspace{0.5cm}
\rotatebox{90}{\hspace{0.1cm} VAE}
\includegraphics[width = 0.95\textwidth]{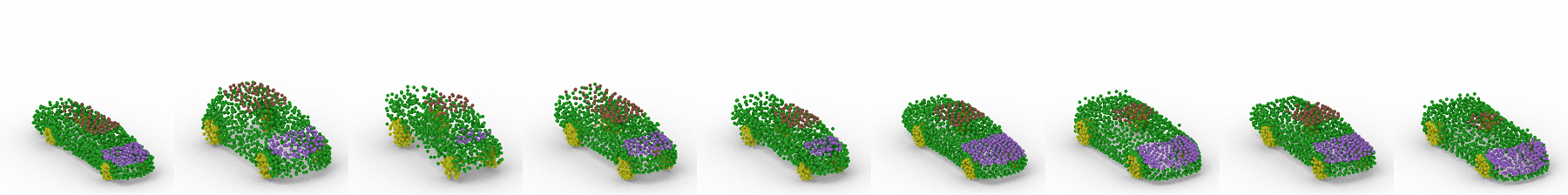}\\
\rotatebox{90}{\hspace{0.1cm} GAN}
\includegraphics[width = 0.95\textwidth]{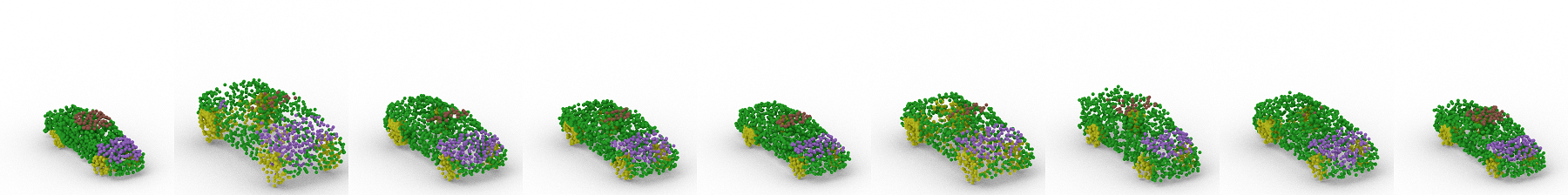}\\
\rotatebox{90}{\hspace{0cm} WGAN}
\includegraphics[width = 0.95\textwidth]{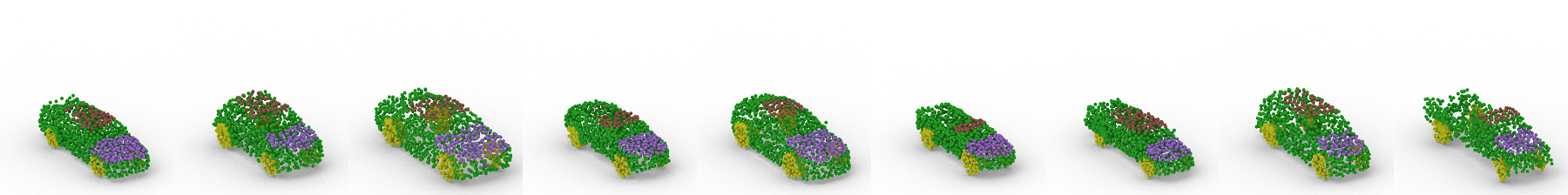}\\
\vspace{0.5cm}
\rotatebox{90}{\hspace{0.2cm} VAE}
\includegraphics[width = 0.95\textwidth]{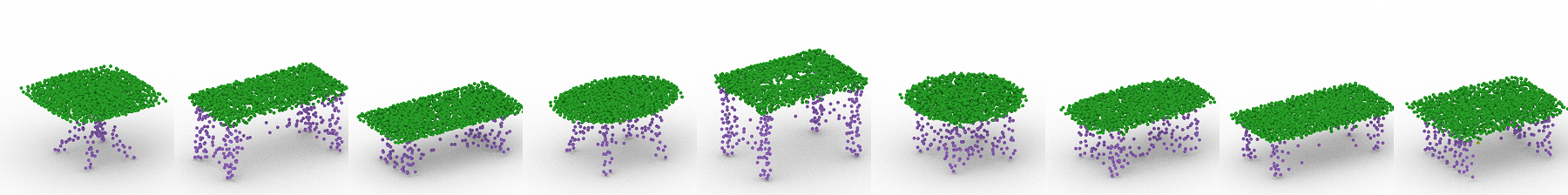}\\
\rotatebox{90}{\hspace{0.2cm} GAN}
\includegraphics[width = 0.95\textwidth]{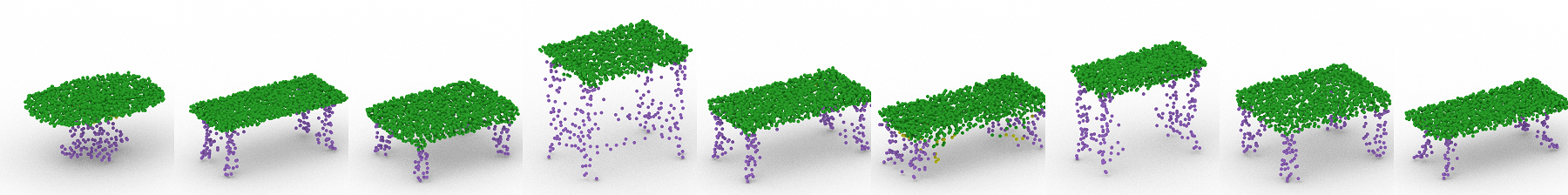}\\
\rotatebox{90}{\hspace{0.1cm} WGAN}
\includegraphics[width = 0.95\textwidth]{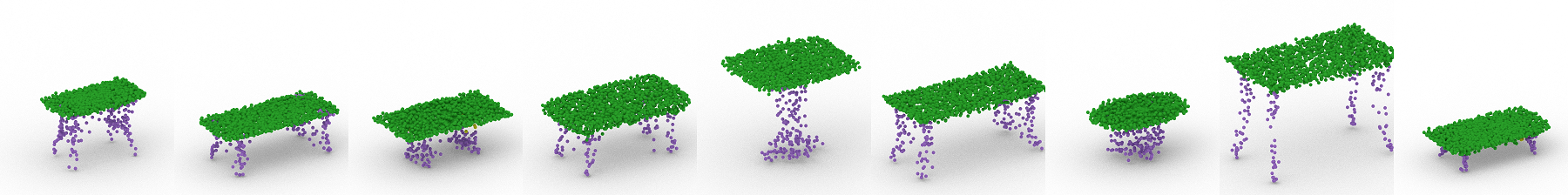}\\
\caption{Samples from generative models for plane, car and table classes.}
\label{fig:apgen}
\end{figure*}

Visualization results for part interpolation and generative models for \textit{plane}, \textit{car} and \textit{table} classes can be found in Fig. \ref{fig:apinterp} and \ref{fig:apgen} respectively. 

\end{document}